*algorithms*

# CONDA-PM—A Systematic Review and Framework for Concept Drift Analysis in Process Mining

Ghada Elkhawaga [1,2,*] , Mervat Abuelkheir [3] , Sherif I. Barakat [2] , Alaa M. Riad [2] and Manfred Reichert [1]

1   Institute for Databases and Information Systems, Ulm University, 89081 Ulm, Germany; manfred.reichert@uni-ulm.de
2   Faculty of Computers and Information, Mansoura University, 35516 Dakahlia Governorate, Egypt; sheiib@mans.edu.eg (S.I.B.); amriad2014@gmail.com (A.M.R.)
3   Faculty of Media Engineering and Technology, German University in Cairo, 11835 New Cairo, Egypt; mervat.abuelkheir@guc.edu.eg
*   Correspondence: ghada.el-khawaga@uni-ulm.de; Tel.: +49-731-5024-131

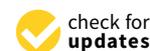

**Abstract:** Business processes evolve over time to adapt to changing business environments. This requires continuous monitoring of business processes to gain insights into whether they conform to the intended design or deviate from it. The situation when a business process changes while being analysed is denoted as Concept Drift. Its analysis is concerned with studying how a business process changes, in terms of detecting and localising changes and studying the effects of the latter. Concept drift analysis is crucial to enable early detection and management of changes, that is, whether to promote a change to become part of an improved process, or to reject the change and make decisions to mitigate its effects. Despite its importance, there exists no comprehensive framework for analysing concept drift types, affected process perspectives, and granularity levels of a business process. This article proposes the CONcept Drift Analysis in Process Mining (CONDA-PM) framework describing phases and requirements of a concept drift analysis approach. CONDA-PM was derived from a Systematic Literature Review (SLR) of current approaches analysing concept drift. We apply the CONDA-PM framework on current approaches to concept drift analysis and evaluate their maturity. Applying CONDA-PM framework highlights areas where research is needed to complement existing efforts.

## 1. Introduction

Driven by the need of continuously improving the performance of business processes as well as the emergence of huge amounts of event data for running processes, process mining techniques and tools have been used for more than a decade [1]. Process mining aims to find a connection between the process model defining how a business process shall be executed and the event log recording data on the actual execution of the process instances by a *Process-Aware Information System (PAIS)*. Process mining aids in gaining insights on actual process behaviour through the analysis of event logs and process models. Although process mining has become a fundamental research area during the last decade, there exist many challenges inherited from those parent research fields. One of the fundamental challenges is *concept drift analysis.*

In response to the evolution of business processes, research on concept drift analysis has gained increasing interest. According to the *process mining manifesto* [1], concept drift refers to *the situation* *in which the process is changing while being analysed*. Research on concept drift addresses the need for monitoring business process changes. Moreover, concept drift allows making a decision about a change, for example, whether to accommodate it or reject the change in case it violates the process specification. Many factors may affect this decision making process. For example, an incremental drift can be easier to trace back and mitigate its effects than a sudden drift. This way, the type of drift may affect the choice whether to propagate the drift and incorporate it as a part of the business process. Another example regards how the granularity level at which the drift occurs may affect the decision, as a drift at the process type level would affect more process instances than a drift at the process instance level.

Ignoring concept drift detection and localisation widens the gap between the intended process and its actual execution. It further complicates the monitoring of running processes and makes results of process analyses outdated. Predicting the future outcomes or activities of a process instance (i.e., predictive monitoring) yields unreliable results if the prediction model is built upon assumptions that may change during process execution. Therefore, it is essential to obtain an understanding of what shall be considered as a concept drift, how to detect a concept drift, and which techniques are available for this detection.

Despite its recognition as one of the major challenges in process mining [1], there is a lack of in-depth research on concept drift analysis and monitoring. As explained in this article, concept drift analysis has several dimensions for example, directions, handling modes, process perspectives, and themes. We argue that there is a lack of studies providing a comprehensive view of the problem along with its solutions. Furthermore, the purpose of this systematic literature review (SLR) is to clarify the confusion of the terms used in current studies and to clarify what is meant by concept drift analysis. We study how current approaches to concept drift analysis contribute to this topic and to what extent the various dimensions of the problem are covered by current research, the SLR discusses the advantages and shortcomings of each approach and systemically compares these approaches. Based on our conclusions from the SLR, we propose the CONDA-PM framework. CONDA-PM is a four-phased framework that may guide process mining practitioners in assessing the maturity level of a concept drift analysis method. CONDA-PM covers a complete lifecycle of a concept drift analysis method with four phases, namely—the *Goals Design*, the *Approach Coding*, the *Implementation*, and the *Evaluation* phases.

Section 1 introduces the contribution of this systematic literature review (SLR). In Section 2, basic terminology is provided, whereas Section 3 introduces different categories of concept drift analysis and distinguishes concept drift from other terms commonly confused with it. In Section 4, we introduce the protocol used to conduct the SLR including research questions, search keywords, study sources, inclusion and exclusion criteria, and results. In Section 5, we analyse the approaches proposed by the studies using the framework to highlight strengths and weaknesses of contemporary research efforts. In Section 6, we build on the findings and introduce the CONDA-PM framework, along with its phases and dimensions. In Section 7, we apply this framework to the approaches used in the selected studies and then evaluate their maturity in respect to concept drift analysis according to a maturity scale we introduce. In Section 8, we discuss our findings and lessons learnt when applying CONDA-PM. Finally, Section 9 highlights related research topics. Section 10 presents a summary and lessons learned.

## 2. Fundamentals

Before defining the research questions pursued in the SLR, some definitions are presented here. These definitions may enable the reader to be informed about fundamental terminology used throughout the rest of the paper. Definitions 1 and 2 are essential to understand the essence of process mining in terms of its inputs and outputs, and hence how different concept drift analysis approaches differ in terms of input artefacts they use. Definition 3 defines different process aspects considered by a concept drift analysis approach. These aspects characterise any business process no matter which representation is used for this process. Definition 4 contrasts two different representations of a business process. One representation considers the effect of time dimension in differentiating resulting process models; whereas the other considers the effect of different execution mechanisms in the differentiation. Definition 5 is essential for understanding a middle artefact used in an approach surveyed in the SLR. Definition 6 is needed to give the reader an introduction about a fundamental artefact that each concept drift approach is relying on (no matter whether it is given as an input or constructed by the approach) to study relations between different activities or events executed in the context of a business process. As a result of studying patterns and inter-activities relations, different versions of the same business process can be compared. Definition 7 provides an introduction to different forms of deriving and representing a business process.

**Definition 1** (Normative and descriptive process models). *A normative process model describes an ideal view on how the activities of a business process shall be performed. It is used for defining how a business process will be executed, and afterwards it is used for monitoring and checking conformance of the running process with the process model. A descriptive model in turn, is specified to capture reality and describe how activities are exactly carried out. In the context of process mining, discovered process models are descriptive.*

**Definition 2** (Event log, event stream). *An event log records the actual execution of a process. Each log entry, that is, each case, corresponds to a specific process instance.*

A case consists of ordered events, each one having one or more attributes, for example, activity name, time, cost, and resource. Each event refers to a single case and may share similarities with other events from other cases, but two events do not have exactly the same values of all associated attributes. Finally, not all events have the same set of attributes, but events referring to the same activity should have the same set of attributes. XES and MXML are well-known formats for representing an event log [2].

Table 1 provides an example of an event log. The case ID column represents the process instance ID. Each case is represented by multiple rows corresponding to the number of events occurred from the start till the completion of this case. Note that in this article, we use the terms *process instance*, *Trace*, and *case* interchangeably to refer to the same concept. However, we tend to use the term *case* whenever we refer to the data view of a business process. Whereas, we use the term *trace* whenever we tend to describe a concept or a related technique concerned with the logical view of a business process. We use the term *process instance* whenever we are concerned with an idea related to the conceptual view of a business process. Each event is recorded by one row with a unique identifier (Event ID column). For any event, it is captured when it occurred (Timestamp column), which activity class it belongs to (activity column), which resource was responsible for this event (Resource column), and which data are associated with it (Cost column).

**Table 1.** A fragment of some event log.

| Case ID | Event ID | Properties | | | |
|---|---|---|---|---|---|
| | | **Timestamp** | **Activity** | **Resource** | **Cost** |
| 1 | 3561 | 30 December 2018 10:02:15 | Enter Loan Application | Sara | 120 |
| | 3562 | 31 December 2018 09:08:22 | Retrieve Applicant Data | Sue | 100 |
| | 3563 | 1 January 2019 12:20:00 | Compute Installments | Mike | 80 |
| | 3564 | 5 January 2019 10:10:28 | Notify Eligibility | Pete | 150 |
| | 3565 | 10 January 2019 08:30:10 | Approve Simple Application | Sue | 50 |

...................

**Table 1.** *Cont.*

| Case ID | Event ID | Properties | | | |
|---|---|---|---|---|---|
| | | Timestamp | Activity | Resource | Cost |
| 7 | 3581 | 30 December 2018 14:02:26 | Enter Loan Application | Pete | 170 |
| | 3582 | 8 January 2019 10:02:15 | Retrieve Applicant Data | Sara | 97 |
| | 3583 | 15 January 2019 15:08:09 | Compute Installments | Sue | 250 |
| | 3584 | 22 January 2019 19:20:10 | Notify Rejection | Pete | 60 |
| | | .................... | | | |
| 55 | 3600 | 30 February 2019 08:44:05 | Enter Loan Application | Mike | 54 |
| | 3601 | 10 March 2019 10:15:25 | Retrieve Applicant Data | Sara | 47 |
| | 3602 | 15 March 2019 11:33:29 | Compute Installments | Pete | 210 |
| | 3603 | 22 March 2019 13:30:44 | Notify Eligibility | Sue | 64 |
| | 3604 | 30 March 2019 09:10:50 | Approve Complex Application | Mike | 120 |

An event stream in turn, corresponds to a collection of events that were created in the context of running process instances. Each event has an embedded identifier to enable matching it with the corresponding activity this event belongs to. In this sense, a log file is a complete record of an event stream. An entry of the log file has a start and an end event, whereas an event stream represents an incomplete running process instance [3].

**Definition 3** (Process perspectives). *The data captured in an event log describe a business process from different perspectives [2]:*

- *Control-flow perspective*, that is, order of activities. Mining this perspective allows characterising and exploring possible execution paths of a process.
- *Organisational perspective*, that is, organisational resources and roles involved in the occurrence of events. Mining this perspective enables the discovery of the social network and the roles participating in a business process.
- *Time perspective*, that is, the point in time events occur. Mining this perspective enables finding bottlenecks and monitoring key performance indicators (KPI) of a business process.
- *Case perspective*, that is, properties of process instances (e.g., values of data elements characterising a process instance). Mining this perspective enables understanding contextual information specific to the current process instance.

**Definition 4** (Process history and process variants). *A process history represents a list of viable process models discovered for one business process [3].*

The process models are discovered at different points in time whenever a violation from the normative process model is detected. The latter is accomplished through checking the conformance of events against the most recent process model from the process history. *A process variant*, in turn, corresponds to *a subset of process instances recorded in an event log*. These process instances have some commonalities (e.g., sharing certain activities with each other), but also show differences to process instances corresponding to other process variants [4].

Process histories are more concerned with the flexibility of a process at the instance level, whereas process variants are studied in the context of business process variability.

**Definition 5** (Execution graph, REGs). *An execution graph is a directed acyclic graph used to represent the execution of a process instance in terms of nodes (i.e., events), edges (i.e., relations between the nodes),* *and functions assigning each event to its activity class [5]. A Representative Execution Graph (REG) corresponds to an aggregation of execution graphs. It is used to pinpoint differences between instances of the same business process [5].*

**Definition 6** (Behavioural profiles)**.** *A behavioural profile is a representation of relations between every pair of events or activities appearing in an event log [6].*

Relations can be causality, direct succession, parallelism, or no-direct succession. A footprint of an event log is one example of a behavioural profile. It represents ordering relations between activities or events in a given event log. These ordering relations can capture certain patterns, like directly-follows ($a > b$), causality ($a \rightarrow b$), choice ($a\#b$), and parallelism ($a||b$). For example, consider an event log L = [$< a, , b, c, d >^3, < a, c, b, d >^2, < a, e, d >$].The footprint for the given log would be similar to the one in Table 2.

**Table 2.** Footprint of event log (L) (adopted from Reference [2]).

|   | a | b | c | d | e |
|---|---|---|---|---|---|
| **a** | $\#L$ | $\rightarrow L$ | $\rightarrow L$ | $\#L$ | $\rightarrow L$ |
| **b** | $\leftarrow L$ | $\#L$ | $||L$ | $\rightarrow L$ | $\#L$ |
| **c** | $\leftarrow L$ | $||L$ | $\#L$ | $\rightarrow L$ | $\#L$ |
| **d** | $\#L$ | $\leftarrow L$ | $\leftarrow L$ | $\#L$ | $\leftarrow L$ |
| **e** | $\leftarrow L$ | $\#L$ | $\#L$ | $\rightarrow L$ | $\#L$ |

**Definition 7** (Procedural and Declarative process models)**.** *procedural process models represent a conceptual view of how activities should be carried out. Meanwhile, declarative process models provide a formalisation of the undesired behaviour through defining a set of constraints.*

Unlike procedural models, the sequence at which activities are executed is not rigid in declarative models [7]. DECLARE is one of the widely used declarative process specification languages [8].

## 3. Concept Drift in Process Mining

Many business environments can be characterised by their turbulent nature and evolutionary changes, at different scales, specializations and velocities. Success of an organisation is judged by its ability to adapt to changes and to incorporate them into its running processes. External factors driving these changes include, for example, changes of business rules, legal regulations, and markets. Changes may also have internal reasons [9], for example, some changes maybe initiated by individuals to adapt to variations in workload or resources; other changes may be ad-hoc. Moreover, organisations may run the same business process, which has been modeled at different levels of granularity with different purposes in mind. As a result, *process variants* represented by models with some common, but also varying activities, emerge. Process variants are different versions of the same process, sharing essential characteristics by conforming to a set of common constraints defined by the process model they adhere to [10]. Another situation holds when process variants result from configurations of the same reference model at build time [11]. Facing different variants of the same process reflects a direct demand of concept drift identification and management.

As a general assumption of most techniques dealing with concept drift, changes may occur in an unexpected and unpredictable manner [12]. In general, changes are either planned [9] or they are introduced spontaneously based on shared knowledge between process participants. As a result, using the normative model to describe a running process would lead to unrealistic analyses, as the process model would describe an idealised version of how work shall be done. Even for descriptive models, the situation would be the same as they would depend on obsolete data from the associated event log. Before presenting the available techniques used to deal with concept drifts, the latter will be defined to avoid any ambiguity.

### 3.1. Distinguishing Concept Drift from Other Concepts

Dealing with concept drift in process mining has its roots in data mining and machine learning, respectively. Concept drift describes *"the situation when the process is changing while it is being analysed"* [1,13]. However, this definition not only neglects the fact that changes may occur at the process type or process instance level, but also considers concept drift as changes at the process type level. However, the latter is not the only level at which changes may occur. Moreover, this view on concept drift ignores the context and reasons of a change, though studying them would ease the drift detection process. Studying the context and reasons of a process change allows identifying whether a drift occurs as part of a corrective decision to a previous problem or to cope with new factors external to the organisation the business process is executed at. As positive side of this definition, the phrase *"while being analysed"*, emphasizes that process miners claim the instability of a business process, that is, it highlights the need for online support and monitoring.

#### 3.1.1. Concept Drift and Noise

The imprecision in the mentioned definition disperses concept drift into different meanings with different purposes and associated techniques. An example of this dispersion is *noise* which refers to *rare and infrequent behaviour* [2]. Noise is considered as a challenge to any process mining approach [1] and is filtered out in several process mining techniques, like heuristic miner [14] and fuzzy miner [15]. A problem in this context is to find a balance between producing underfitting and overfitting models. Although noise is allowed according to the process model, it is not preferred to be taken into account when discovering a descriptive model, as its inclusion would result in unstructured spaghetti-like models.

#### 3.1.2. Concept Drift and Deviations

Referring back to noise, deviations and deviance mining have different concerns. *deviations* are defined to be *"additional behaviour observed in the event log, but not foreseen in the normative process"* [5]. Aligned with this definition, the authors of [16] denote *deviance mining* as *"a family of process mining techniques aimed at analyzing event logs in order to explain the reasons why a business process deviates from its normal or expected execution"*. These deviations include violations of compliance rules or the exceeding or undershooting performance targets [16]. As argued by the authors in [2], however, having a dependable normative model for a process constitutes a difficult and error-prone task, as models are expressed at various abstraction levels and capturing human behaviour through a model is very difficult. The authors of [5] argue that available techniques for deviance mining classify traces as either deviant or normal, regardless of their control-flow execution, but rather based on performance-related measures (e.g., throughput time). By contrast, the authors of [16] show that the input of deviance mining should be traces labeled as either *normal* or *deviant*.

#### 3.1.3. Concept Drift and Process Drift

In [9], the authors define *business process drift detection* as *"a family of techniques to analyse event logs or event streams generated during the execution of a business process in order to detect points in time when the behaviour of recent executions of the process differs significantly from that of older cases"*. This definition limits drift detection to be solely based on event logs, while ignoring the efforts exerted in the areas of graph matching and process variants analysis. Concept drift analysis can be based either solely on the process model or the process model as well as its corresponding event log data. As example, consider *delta analysis*, which can be defined as a comparison of normative and descriptive models representing a certain process to check for their similarity or disparity [17]. Another artefact that may be used in the context of concept drift analysis is the *change log*. The latter is created and maintained by, for example, adaptive PAISs, which maintain change logs containing information about process changes at both type and instance level [18]. In this context, the authors of [18] introduce change mining, which uses change logs to obtain an abstract change process reflecting the changes applied to a business process either on the type or instance level.

### 3.1.4. Variability and Flexibility

Although BPM practitioners always push towards a flexible business process, flexibility levels are not well defined. According to [19], flexibility is *"the ability to react to external changes by modifying only those parts of a process that need to be changed while keeping other parts stable"*. However, this definition distinguishes neither between internal and external changes, nor between the levels at which the changes take place. By the latter we mean whether a change is introduced at the process type level and then propagated to selected traces, or comes bottom-up from the traces to the process level. The authors of [20] distinguish between *variability*, which corresponds to the process of providing a customisable process model, and *flexibility* which regards the changing circumstances and variations at the process instance level. Variants of a business process should adhere to the customisable process model. However, enabling variability by having a customisable process model is different from the notion of reference model, as the latter provides a blueprint rather than it enables customisation at the structural level of a model [20]. Variability governs and organises the high-level change process, while flexibility organises the low-level change process at the instance level.

### 3.2. Categories of Concept Drift Analysis

To explore concept drift analysis process, its dimensions need to be thoroughly explored. These dimensions can be broken down into types of concept drift, and concept drift analysis tasks.

### 3.2.1. Concept Drift Types

An important component of concept drift analysis process is realising that concept drift can take different forms. As will be illustrated through the SLR and CONDA-PM, choosing the right concept drift analysis technique is correlated with the type of concept drift addressed. Choosing unsuitable technique can result in ignoring important patterns in the data that may result in detecting a different type of the drift other than the one present in the data under analysis.

Figure 1 illustrates these different concept drift patterns. Regarding *sudden concept drift*, a change takes place dynamically during the execution of a process instance. A sudden drift may affect the way or the point in time an activity may be executed, the resources performing the activity, or any other process perspective concerned by a change. As example on a sudden drift consider the change of the process of uploading a video to any media streaming website to ensure the ownership of this media as a consequence of the European copyright legislation.

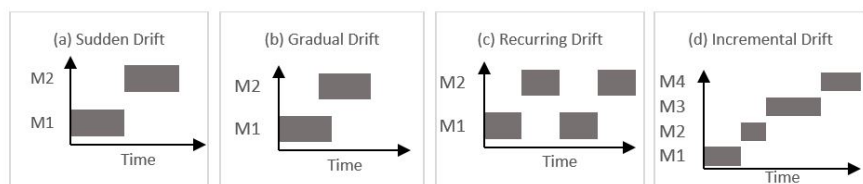

**Figure 1.** Different types of concept drifts (adopted from Reference [13]).

*Gradual concept drift*, in turn, is characterised by the co-existence of two or more versions of the process; that is, the old process version exists along with the new one, over a certain period of time, and hence, process instances may be executed according to the old or the new version. Afterwards, the new version operates solely, that is, the process instances may be only created from the new version. For example, an organisation O1 from the field of consultation services is acquired by an organisation O2 adopting another customer service process model M2. The process model of the acquiring organisation M2 will be applicable only to new customers of O1, whereas the current consultation traces that have not been completed before the acquisition follow the customer service model of O1. After completing all current traces, process model M1 of O1 is no longer used.

*Recurring concept drift* can be defined as a drift taking place over a certain period of time. Afterwards, the process switches back to its former version. This drift is caused by the process context or environment. As example, consider occasional changes of a marketing process during a certain period of the year, for example, to offer reductions for certain products during public holidays based on the country or the state of the customer. Note that the duration of the offer and the period it is applied are both subject to external factors.

*Incremental concept drift* takes place when a change is introduced incrementally into the running process until the process reaches the desired version. An example of incremental drift is the incremental introduction of self-service terminals for paying fines at toll stations [8].

Figure 2 illustrates the fifth type of concept drift, as introduced in Reference [21], namely *multi-order drift*. The latter involves changes on two granularity levels, that is, micro and macro level. As illustrated in Figure 2, M1 and M2 represent the macro level of the change to be introduced to a business process; that is, M1 represents the current process an organisation adopts and M2 corresponds to a new process to be adopted over a time period of, for example, 12 months. M1 is further divided into M11 and M12, and M2 into M21 and M22, respectively. The resulting divisions of M1 and M2 represent the micro level of the change; they allow introducing the total change as chunks of different periods of time, rather than introducing M1/M2 suddenly. When M1 becomes fully adopted, M11 and M12 will recur, each through a period of 6 weeks. After 6 months, M2 will be adopted and its variants (i.e., M21 and M22) will recur every 6 weeks.

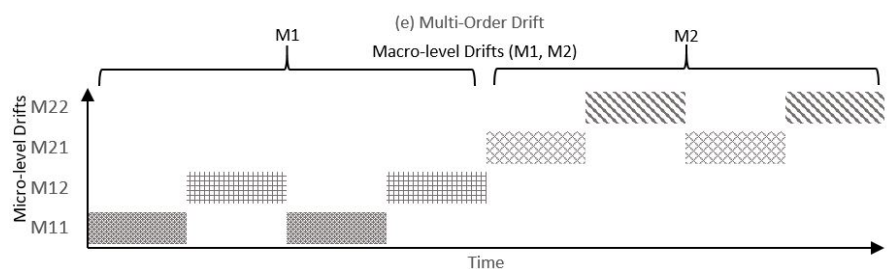

**Figure 2.** Multi-Order drift adopted from ([21]).

Regarding this type of concept drift, different population sizes (i.e., volumes of traces) are considered as well as different time scales. Studying change at different time scales allows switching between the detection of micro level changes, when considering shorter time scales, and the detection of macro level changes, when considering larger time scales. For example, assume that an organisation O1 is acquired by organisation O2. Both have different invoicing models for their products. Oragnisation O1 uses a 3-way matching model M1 with two versions, *invoice after goods receipt* M11 and *invoice before goods receipt* M12. Oragnisation O2 uses another invoicing model M2 with two versions, that is, a 2-way matching model M21 and a consignment model M22. O1 is switching between M11 and M12 based on the product type. O1 shall fully adopt M2 after 6 months. After adopting M2, O1 will use both M21 and M22 and switch between them based on the product type.

### 3.2.2. Concept Drift Analysis Tasks

In [13], the authors provide a characterisation of concept drift in terms of the tasks required to handle it:

1. Detecting the point of time when a drift took place. This task is about detecting a drift as well as the time period at which the drift occurred.

2. Localising the part of the process where the drift happened and characterising the process perspective affected by the drift, for example, the control–flow, data, or resource perspectives. Moreover, this category includes identifying the pattern of the concept drift, that is, whether the drift is sudden, gradual, incremental, recurring, or multi-order.

3. Discovering the evolution process. This task involves putting the detected drift into a larger context, in order to discover how the business process is changing. Discovering the change process can be jointly achieved by studying the complete set of detected drifts for this process, rather than analysing the drifts in isolation from each other.

As will be illustrated though the SLR and CONDA-PM, these tasks are not necessarily combined in one approach. However, they are related in a hierarchical form. Concept drift analysis tasks can be accomplished *offline*, after completing an instance, or *online* during the execution of an instance. In [22], empirically evidenced process change patterns are presented, including insertion, movement, deletion, and replacement or activities and process model fragments, respectively. Most process change patterns refer to the behavioural perspective (i.e., control flow). In general, change patterns are useful to derive high-level changes from low-level change primitives. Moreover, the authors in [22] identify another dimension of a process change, concerning its duration (i.e., whether this change is momentary or permanent). This dimension is critical to make a decision on whether to propagate a change to the process type level or just keep it solely at the process instance level.

### 3.2.3. Putting all the Pieces Together

In Reference [19], a generalization of these categories is presented by abstracting process flexibility classifications into three main aspects:

- Where do changes occur (i.e., abstraction level of the change)?—This aspect focuses on whether a change occurs at the process type or the process instance level.
- What is being changed (i.e., subject of change)?—This aspect concerns the process perspectives affected by a change.
- How to characterise a change (i.e., properties of change)?—This aspect can be mapped onto the drift patterns and duration of a drift?

Regarding the abstraction level of a process change, the authors of [20] distinguish between flexibility and variability. Moreover, the work in [20] introduces the concept of *flexibility by change* which defines a condition where a certain change is applied to selected traces at runtime. In turn, this might qualify the change to be leveraged to the process type level, that is, to be propagated to all process instances following the same process variant. Furthermore, the authors of [20] introduces *variability by restriction*, which starts with providing a process model with all allowed behavior and restricting models that may be configured from this model by adding more behavior. On the contrary, *variability by extension* begins with providing a process model whose behaviour can be extended by deriving other process models based on the behaviour explained by it. In general, mining flexible processes is challenging as the employed business process might be rather unstructured, resulting in spaghetti-like models. The authors of [23] describe this situation by having process models with a large number of nodes and interrelations resulting from the diversity of an event log.

Another effort to characterise the levels of concept drift is introduced in the work in [24], which considers concept drift as the action of changing the associations between elements, labels, or references. Any change of the associations between elements is considered to be concept drift, whereas changes of labels and references are considered to be a conceptual replacement. Adopting this view, a concept drift would occur whenever a change affects the ordering relations (e.g., directly follows, parallel, causal dependency, choice) between a set of activities. Adhering to the same view, conceptual replacement takes place whenever a change happens to the name of an activity or the actual task represented by this activity. This viewpoint can directly relate to Semantic Web, and to some extent to process mining, as well.

## 4. Research Methodology

The systematic literature review we conducted in the context of this work was designed according to the guidelines suggested by Kitchenham & Charters [25]. The SLR protocol is presented in the following subsections. Applying it ensures a reproducible and accurate scientific contribution, and actively reduces any bias regarding the collection and analysis of the studies upon which CONDA-PM (cf. Section 6) is derived.

### 4.1. Research Questions

The aim of the SLR is to gain an in-depth understanding of the solutions proposed by current approaches to concept drift analysis in the context of process mining. We are concerned with reviewing studies on concept drift analysis in order to derive characteristic properties of corresponding analysis approaches in terms of inputs, processing, and outputs. These properties are used afterwards for developing an understanding of what constitutes a comprehensive approach to concept drift analysis in process mining. This understanding, in turn, provides the basis for deriving the CONDA-PM framework we introduce in Section 6. In detail, the SLR investigates the following research questions:

1. RQs investigating the inputs of an approach:

   - *RQ1: What are the inputs needed by a concept drift analysis approach?*—This question targets identifying the artefacts used by an approach as starting point for concept drift analysis. This kind of input affects the choice of the techniques to be used at the presence of this input, the expected outcome, and the kind of change to be studied, that is, whether the approach is concerned with analysing behavioural or structural process change.

2. RQs investigating the processing phase of an approach:

   - *RQ2: How can concept drift analysis be classified to gain a more profound understanding of the characteristics of the various approaches?*—Trying to find some matching criteria between approaches may help in gaining understanding of the potential building blocks of a concept drift approach, and hence, enabling forming this understanding into a guiding framework. Finding and exploring differences between contemporary approaches may help in identifying specific application scenarios for these approaches.
   - *RQ3: Which process perspectives are subject of concept drift analysis processes?*—This research question studies which perspectives of an event log are covered an existing approach, which is crucial to judge the expressiveness of the approach. For example, the control-flow perspective may be used to study both structural and behavioural changes of a business process, whereas studying changes concerning the time perspective can facilitate the task of predictive monitoring, enriching it with information beneficial for performance-related predictions.
   - *RQ4: Which granularity level of a business process is mainly addressed by existing concept drift analysis approaches?*—Studying the granularity level of a change (i.e., whether the change is applied at the process type or process instance level) is crucial to derive insights on the prevalence and effects of the change. A drift on the process type affects the related process variants, and hence, a larger number of process instances than a drift on the instance level which is limited to one single variant of this business process. Identifying the level at which a change occurs is important in making a decision on how to handle the change and in calculating the costs of making corrective actions or propagating the change.

3. RQs investigating the outputs and evaluation of an approach:

   - *RQ5: Do existing approaches provide usable implementations?*—The maturity of any approach is reflected by the availability of a technical implementation, being able to demonstrate its applicability and sustainability.

- *RQ6: How does an approach to concept drift analysis foster a business process proactivity to change?*—Studying a concept drift in an online setting has different implications than studying it in an offline mode. Furthermore, the ability of an approach to precisely define both the moment and type of drift is crucial for guiding the decision making process. Yielding precise outcomes of a concept drift analysis approach affects related procedures, for example, predicting the outcome of a process instance when knowing that this process instance was subject to a drift during its execution. Having accurate information about a drift enables process practitioners not only to react to changes, but also to make decisions about future improvements of the business process. Process practitioners may also make decisions taking future drifts into account. This way, concept drift analysis outcomes enable business process proactivity to changes.
- *RQ7: How to evaluate concept drift analysis approaches?*—Studying the nature and types of event logs contemporarily used to evaluate existing approaches is crucial for any successful evaluation process. In particular, three factors are crucial—the artefacts used for the evaluation, the techniques used, and the questions asked. For example, using an artificial event log tailored towards a specific scenario affects the ability to generalise the approach under evaluation. Note that the maturity of an approach is highlighted through the evaluation phase.

Note that the research questions are closely related. Based on the type of the input processed by a concept drift approach studying concept drift, a certain technique can be applied and different kinds of changes be studied. Consequently, it is important to discover this relation through the results of the SLR.

The SLR is concerned with identifying which inputs current approaches use for analyzing concept drift. There is a need to define whether these inputs are made available through PAISs or are generated in a pre-processing step. As another goal of the SLR, we want to categorize concept drift-related approaches. The resulting categories shall shed light on the techniques used for analysing concept drift and the SLR shall provide a means to study the availability of related tools and the relation between the used techniques and the possibility of implementing tools supporting these approaches. The SLR further aims to explore which process perspectives are mostly covered by existing approaches to concept drift analysis. Finally, the SLR aims to identify whether there is a lack of research studying concept drift in one or more of these event log perspectives. This helps associating concept drift analysis with other open issues like predictive monitoring, as drifts in perspectives like time are explored. The granularity level at which a concept drift is studied is further needed to place measurements of the business process robustness or degree of agility, and hence, to take the suitable decisions, for example, replacing the whole process model.

### 4.2. Search Keywords

To answer the research questions, we defined different combinations of search strings. The search keywords were sometimes more general to broaden the search area, and often more specific and precise. The search string was elaborated many times as we have gained more in-depth knowledge into the topic. The general keywords resulted in studies that reside in the intersection areas between process mining and data mining as well as process mining and business process management. We excluded terms that yielded studies from unrelated fields, for example, mining, or fields not directly concerned with concept drift analysis, for example, predictive monitoring. Finally, the search strings we used were as follows:

"Concept drift in process mining" OR "change mining" OR "concept drift in business process" OR "concept drift in event log" OR "change in business process" OR "deviation in process mining" OR "deviance mining in business process" OR "business process variants" OR "mining process variants"

The search string was refined several times to obtain more relevant studies according to the criteria defined in the following subsections.

*4.3. Sources*

The search string was applied to *five electronic libraries* in order to discover studies related to the topic "concept drift". These electronic libraries are as follows:

- *Science Direct - Elsevier* (https://www.sciencedirect.com/)
- *IEEE Xplore Digital Library* (https://ieeexplore.ieee.org/Xplore/home.jsp)
- *SpringerLink* (https://link.springer.com/)
- *ACM Digital Library* (https://dl.acm.org/)
- *Google Scholar* (https://scholar.google.com/)

These libraries contain repositories of journal articles as well as the most relevant conference and workshop proceedings in the field. Furthermore, the libraries provide capabilities of filtering results based on different criteria, like year of publication, subject, and type (e.g., conference proceedings, journal article, technical report). Moreover, Google Scholar alerts were set to trigger us when recent studies on the topic, which match parts of the search string, had emerged. Based on this variety of libraries, we aimed to retrieve high quality published studies with a minimum overlap between them. Finally, we applied *backward reference searching* by further examining sources referenced by more than one of the included studies. The results were saved in an Excel sheet (https://drive.google.com/file/d/1KLdtI5yg4EiN_Fw-AMEgWMuIJjDuIDLA/view?usp=sharing), to which inclusion and exclusion criteria were applied. Each study was initially analysed to extract the following information:

- *Abstract.* This criterion was the necessary for gaining an overview of the topic and contributions of a study.
- *Title.* This criterion was the determinant in the first round of selecting relevant studies to be included in the SLR.
- *Year of publication.* The more recent a study is, the more it builds upon knowledge gained from previous works. Furthermore, more recent publications constitute potential sources for identifying other studies by conducting backward reference searching. Finally, less recent studies have been expected to be analysed in more recent ones, that is, this kind of analysis was surveyed as well.
- *Type.* This criterion was one of the determinants of the maturity level of a study, for example, publications in conference proceedings are expected to provide more mature approaches than the ones published in workshop proceedings. Moreover, journal publications are considered being most mature and detailed.
- *Publisher.* The publishing institution has a weight on the expected quality of research done through a study. For example, publishers like Elsevier, IEEE Xplorer, and Springer are known for editing high impact factor journals and conference proceedings.
- *Number of citations.* This criterion was one of the indicators of the relevance on the considered approach. Sometimes, the number of citations was influenced by the year of publication. Furthermore, we came to the conclusion that the number of citations is influenced by the direction an approach is pursuing. This intuition was developed after deriving the directions of studies, which are introduced in Section 5. However, we did not consider the number of citations as a main criterion for including or excluding a study as it will be affected by the year of publications, especially for the studies published between 2017 and 2019.

*4.4. Inclusion and Exclusion Criteria*

To restrict the studies to the most relevant ones, we defined the following inclusion and exclusion criteria.

- Inclusion criteria:

1. The study uses techniques relevant for analyzing concept drift in the context of process mining.
2. The study deals with a case study that is relevant to process mining and there is a possibility of concept drift in this case study.
3. The study uses event log-related techniques for analyzing concept drift.
4. The study deals with change analysis in business processes.

- Exclusion criteria:

1. The study deals with a topic in process mining other than concept drift.
2. The study deals with concept drift in a field other than process mining.
3. The study is a technical report or a thesis.
4. The study is not presented entirely in English language.
5. The study is not publicly available or is not published in conference proceedings, journals or books.

*4.5. Study Selection*

The SLR construction process started with defining the SLR goals by shaping the pursued research questions (cf Figure 3, Stage 1). Identifying the relevant studies was based on the presented search string (cf. Section 4.2) over the five search libraries (cf. Figure 3, Stage 2), resulted in a total of 292 studies. This result set was filtered by applying the inclusion criteria (cf. Section 4.4), reading both the title and abstract of each study to ensure its relevance to the SLR (cf. Figure 3, Stage 3). Afterwards, we applied the exclusion criteria (cf. Section 4.4), which resulted in 44 studies. These studies were input to the backward reference search process (cf. Figure 3, Stage 4). Then inclusion and exclusion criteria were applied again, and the final result comprised of 19 studies. Furthermore, the selected studies were categorised (cf. Figure 3, Stage 5) and the result is four categories (cf. Section 5). The selected studies are presented in Table 3. Each study has a unique ID composed of the letter 'S' concatenated with a number. Note that interest in the topic has increased over the years, with a peak in 2017. However, a decrease in the number of studies can be observed in 2018. This decrease can be explained by the growing trend of integrating research in concept drift with other process mining topics like predictive monitoring.

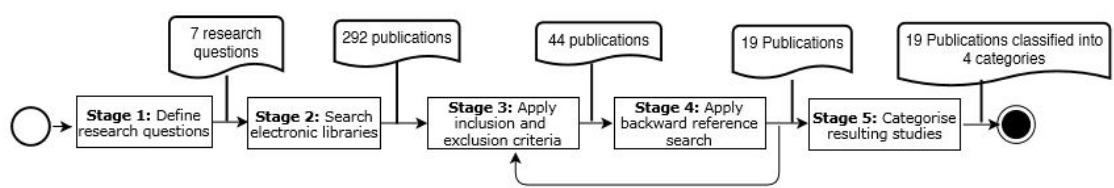

**Figure 3.** Stages of study selection and analysis process.

For each research question, answers are extracted from the selected studies, which are highlighted by green color in the excel sheet mentioned in Section 4.3.

When analysing selected studies, another observation may be made regarding the type of study. We found that 12 studies are conference publications, three are published in a workshop, only two are journal publications, and two are symposium publications . This observation can be explained by the fast feedback cycle an author obtains through conference publications. However, it may have a serious indication regarding relevant research maturity in the topic of concept drift analysis.

**Table 3.** Relevant studies resulting from the search process (descending by publication year).

| Study Id | Authors | Year | Publication Type | Bibliography No. |
|---|---|---|---|---|
| S1 | Stertz & Rinderle-Ma | 2019 | Conference | [26] |
| S2 | Yeshchenko et al. | 2019 | Conference | [8] |
| S3 | Stertz & Rinderle-Ma | 2018 | Conference | [3] |
| S4 | Nguyen et al. 2018 | 2018 | Conference | [27] |
| S5 | Maaradji et al. | 2017 | Journal | [9] |
| S6 | Hompes et al. | 2017 | Symposium | [28] |
| S7 | Bolt 2017 | 2017 | Conference | [4] |
| S8 | Zheng et al. | 2017 | Conference | [29] |
| S9 | Seeliger et al. | 2017 | Conference | [30] |
| S10 | Lu et al. | 2016 | Workshop | [5] |
| S11 | Ostovar et al. | 2016 | Conference | [31] |
| S12 | Martjushev et al. | 2015 | Conference | [21] |
| S13 | Hompes et al. | 2015 | Conference | [32] |
| S14 | Bose et al. | 2014 | Journal | [13] |
| S15 | Buijs & Reijers | 2014 | Conference | [33] |
| S16 | Van der Aalst | 2012 | Conference | [34] |
| S17 | Carmona & Gavaldà | 2012 | Symposium | [35] |
| S18 | Luengo & Sepúlveda | 2012 | Workshop | [36] |
| S19 | Song et al. | 2009 | Workshop | [23] |

### 4.6. Data Extraction

For each of the 19 studies, a *data extraction strategy* is applied with the aim to answer the research questions defined in Section 4.1. After analysing and splitting each RQ into parts, we extract the following study aspects from each selected study:

1. *Inputs.* This aspect deals with the necessary inputs used by the approach proposed in the study. An approach may have one or multiple inputs of different types (cf. RQ1).
2. *Change form.* This aspect indicates whether the approach tries to identify drifts in the form of behavioural or structural changes (cf. RQ6).
3. *Regarded perspective.* This aspect indicates which process perspectives a study considers, that is, whether to identify drifts regarding process activities and their control flow, or other aspects like time or organisational resources. The more perspectives a study covers, the more comprehensive it is (cf. RQ3).
4. *Study focus.* This aspect describes which concept drift analysis task the study is concerned with (cf. RQ6).
5. *Handling mode.* This aspect indicates whether the tasks of concept drift analysis are accomplished online or offline by the approach proposed in the study (cf. RQ6).
6. *Drift patterns.* This aspect represents the type of drift is addressed by the approach (cf. RQ6).
7. *Theme.* This aspect shows whether the approach represents an attempt to analyse concept drift at the process type level indicating the (variability) of a process, or at the process instance level referring to the (flexibility) of this process (cf. RQ4).
8. *Used techniques.* This point indicates the category of techniques used by the approach (cf. RQ2).
9. *Tool availability.* This point indicates whether there is an implementation of the approach (cf. RQ5).
10. *Evaluation input.* This point represents the types of inputs the study relied on when evaluating the proposed approach (cf. RQ7).
11. *Evaluation (What/How).* This point shows how the approach was evaluated (cf. RQ7).

Table 4 summarises the data extraction process and shows how the derived study aspects were mapped to answer the proposed RQs. In particular, for each criterion, Table 4 describes the type of data extracted from each study, as well as the analysis type, that is, how this data is extracted.

The data extracted with respect to the first five criteria (of Table 4) are saved in the Excel sheet, collecting the data of all studies (cf. Section 4.3). These data are not further processed. To foster readability of Table 4, we shortly explain the values of the entries in columns *Extracted data* and *Analysis type*:

- *Initial list based on previous knowledge*. This corresponds to a list constructed by the first author of this paper and revised by one of the co-authors. This list is based on previous knowledge of the point and its possible values in the context of process mining.
- *Predefined list*. The values of this list are extracted from either an existing cited study, or common preliminary knowledge of concept drift and process mining.
- *Frequency counts*. This count is used when the value of the criterion is clearly stated in the study.
- *Content analysis techniques*. Techniques described in Reference [37] to infer and extract information from text are applied.

**Table 4.** Data extraction results.

| RQ | Criteria | Extracted Data | Analysis Type |
|---|---|---|---|
| General Information | Title | Free text | None |
| | Year | Date | None |
| | Type | Free text | None |
| | Publisher | Free text | None |
| | Number of citations | Number | None |
| RQ1 | Inputs | Initial list based on previous knowledge | Frequency counts |
| RQ5 | Change form | Initial list based on previous knowledge | Content analysis techniques |
| RQ4 | Regarded perspectives | Predefined list based obtained from [2] | Frequency counts |
| RQ5 | Study focus | Predefined list based on existing concept drift task | Content analysis techniques |
| RQ5 | Handling mode | Predefined list obtained from [22] | Frequency counts |
| RQ5 | Drift patterns | Predefined list based on existing concept drift types | Frequency counts |
| RQ6 | Theme | Predefined list obtained from [20] | Content analysis techniques |
| RQ2 | Used techniques | Free text | Content analysis techniques |
| RQ3 | Tool availability | Free text | None |
| RQ7 | Evaluation input | Numbers, free text | None |
| RQ7 | Evaluation (What/How) | Free text | Content analysis techniques |

## 4.7. Highlights after Data Extraction

Studies investigated through the data extraction process reflect important conclusions that can highlight the shortcomings of the respective research dealing with concept drift in process mining. We aim to reach an understanding of the goals of existing approaches. Thus we begin with studying how the selected studies are distributed among different handling modes (RQ6) and themes (RQ4). According to Figure 4, 15 studies address concept drift in an offline mode, while only four studies enable online concept drift analysis (RQ6). This observation raises an issue concerning the way the existing approaches notify users about potential drifts instead of solely dealing with drifts in a reactive mode. Moreover, it is of interest how existing approaches can yield reliable results that may be used as a basis for complementary tasks like making predictions for a running process instance, while the underlying data is experiencing a concept drift.

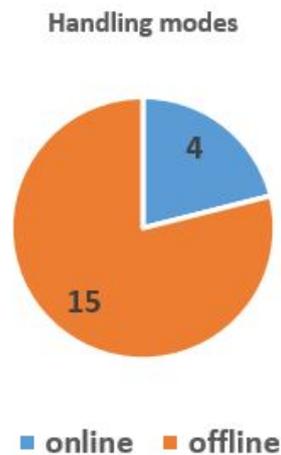

**Figure 4.** Number of publications classified according to handling modes.

Figure 5 shows that 14 studies deal with the flexibility of a business process at the process instance level, whereas four studies are concerned with studying concept drift at the process type level, and only one study is addressing both. This can be explained with the fact that when moving higher in the granularity level of concept drift analysis (i.e., when analysing concept drift at the process type level), different other topics (e.g., process variability, change mining) emerge whose enabling techniques may interleave with concept drift. As a result, the added value and purpose behind concept drift analysis might be missing.

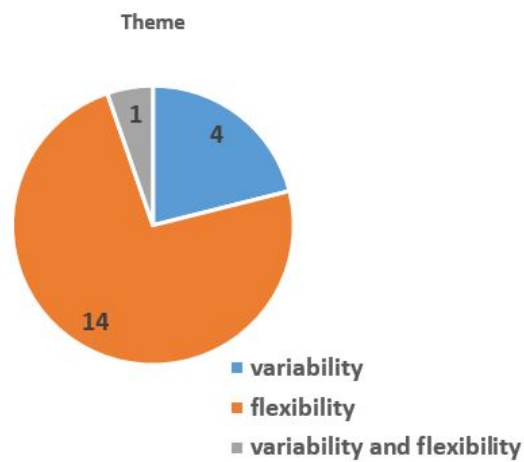

**Figure 5.** Number of publications according to themes of handling.

When studying the inputs needed by an approach to analyse concept drift and the way the variance in these inputs might affect the focus of an approach (RQ1), it can be noticed that all selected studies take an event log as input. Note that only three studies (S9, S15, and S16) take the process model as an additional input. Figure 6 shows the number of studies considering one or more process perspectives (RQ3). 12 studies solely deal with changes in the control flow perspective, while three studies consider two and four studies more than two perspectives. From this, we can infer some factors contributing to that the difference in studied process perspectives. The first is the data available in an event log, that is, whether this event log is solely recording control-flow data or also time, resources, and other data items. Another factor is the used technique and its ability to encode a process instance execution sequence with the associated payload, for example, timestamps and resources.

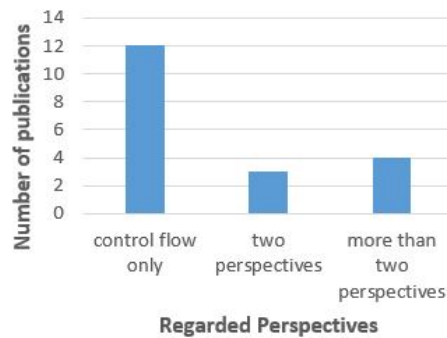

**Figure 6.** Distribution of publications over regarded perspectives.

As an attempt to understand the types of concept drift mainly addressed by selected studies (RQ6), we plot the number of publications with respect to the concept drift types they address (cf. Figures 7 and 8). First, we are interested in whether or not the type of concept drift is stated clearly by the selected study. As can be observed from Figure 7, 11 studies do not explicitly clarify the type of concept drift that can be detected with the proposed approach. As a consequence, issues are raised regarding the clarity and maturity of an approach. However, note that the type of drift is subject to the domain an event log represents as well as to the length of the time period the event log was recorded. From Figure 8, we can observe that sudden drifts are addressed by eight studies, which gives this drift type the biggest share. Recurring and incremental drifts are addressed by three studies (S1, S2, S3) along with sudden and gradual drifts. This can be explained with the fact that S1 is building upon and adding to the work introduced in S3. In turn, multi-order drifts are solely addressed by S12 along with sudden and gradual drifts.

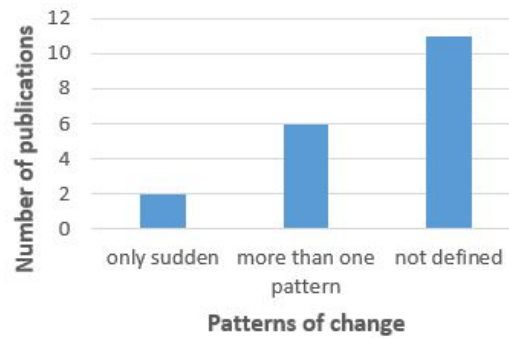

**Figure 7.** Number of publications classified by whether or not concept drift type is stated.

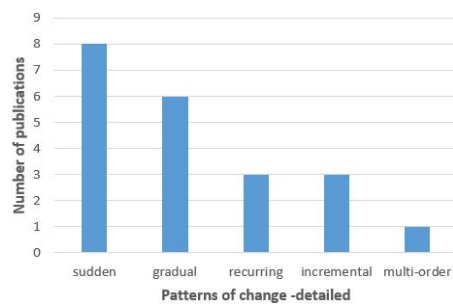

**Figure 8.** Number of publications classified by different types of concept drift.

The core of any approach is represented by the algorithm or techniques it provides to process its inputs such that the desired outcomes can be obtained. When tracing back the relation between inputs, processing, and outcomes, therefore, we visualise the concept drift analysis tasks being of interest to the selected studies and show which change forms are mostly addressed by the studies. We plot the distribution of selected studies over change forms (RQ6), indicating concept drift tasks. As shown in Figure 9, 17 studies address behavioural changes, whereas only one deals with structural changes. One study addresses both forms. This observation complies with the previous observation on the types of inputs used by the various approaches (16 studies take only event logs as input, whereas, three studies take process models into account as well). S9 is one of the three studies that take both an event log and a process model as input, and use the available process model for analysing structural changes.

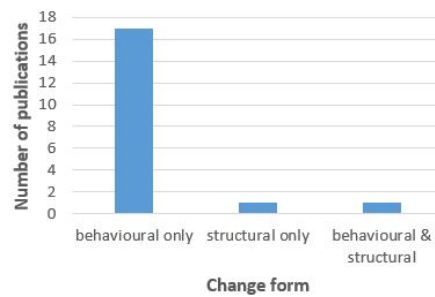

**Figure 9.** Number of publications classified by studied change form.

Figure 10 allows for several observations. *First*, the majority of studies (i.e., nine studies) dedicate efforts for detecting and localising concept drifts. This can be explained with the fact that the use of machine learning algorithms and probabilistic techniques facilitates these tasks if the input (i.e., event log) is rich with information about the state of an activity on a vertical spectrum (i.e., at the current point in time there is enough payload enabling inference about the activity) as well as on a horizontal spectrum (i.e., there are enough process instances that can be used to study the change in the state of an activity). *Second*, four studies either directly or indirectly aim at investigating the evolution of drifts and putting a detected drift into the context of the whole process. This trend allows studying the effect of a drift on the process and enables different other tasks to be accomplished accurately, for example, making more accurate predictions about the outcome or next activity of a running process instance.

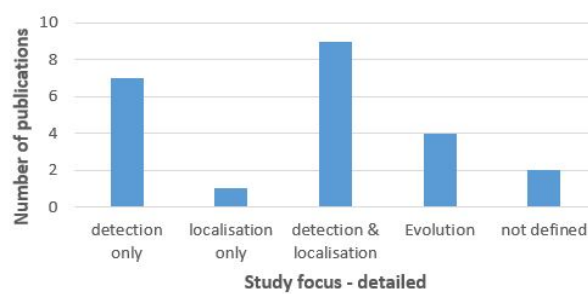

**Figure 10.** Number of publications classified by study focus.

From a practical perspective, it is also important to have insights into the platforms used for implementing concept drift analysis approaches (RQ3) as well as the artefacts used for evaluating the latter (RQ7). Figure 11 plots the selected studies over different platforms. As can be observed, 47% of the studies (i.e., nine studies) use ProM [38] as a platform. This can be explained with the fact that ProM is an open source testbed for all process mining-related approaches. Moreover, some of the approaches rely on other ProM packages for preprocessing the used event logs. A big share of the studies either implement their approach as proof-of-concept prototypes (five studies) or do not provide any implementation of the approach (three studies). Figure 12 plots the artefacts used for evaluating the approach. 52% of them (i.e., ten studies) use both an artificial event log for preliminary experiments and real-life logs for demonstrating the capabilities of the proposed approaches. Real-life logs [39–46] are possibly available from Business Process Intelligence Challenges (BPIC) in the years 2011 to 2019. Furthermore, among the factors contributing to real-life logs availability is the increased awareness of business entities about the importance of process mining and the inherent benefits of its application, for example, gaining more accurate insights into real-life business processes and tuning these processes accordingly.

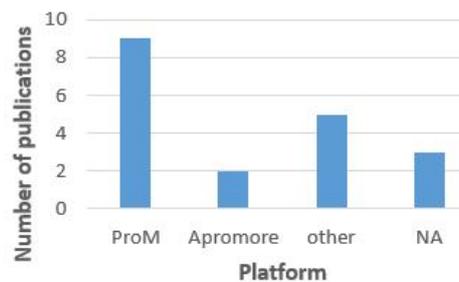

**Figure 11.** Number of publications classified by implementation platform.

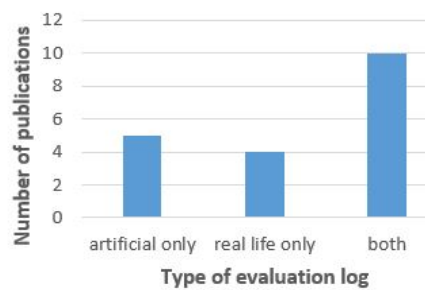

**Figure 12.** Number of publications classified by evaluation artefact.

## 5. Concept Drift Analysis Approaches in Process Mining

Despite the number of proposals and contributions made in the context of concept drift in process mining, there is no comprehensive approach addressing concept drift analysis from different aspects and with all of its types. After surveying literature in process mining in general and concept drift specifically, we notice that research efforts on concept drift analysis have a similar purpose but differ in how to achieve it. We conclude that some of the approaches share commonalities regarding inputs and used techniques. In particular, these commonalities affect the studied change form as well as the process perspectives covered by an approach. This is illustrated in the following subsections that shall answer RQ2.

As a commonality, the presented approaches rely on the presence of an event log. Moreover, they all focus on detecting the drift through the effect it has on the resulting process instances, that is, cases in the event log, rather than detecting it by comparing process models representing different process instances before and after the drift occurs. We present an attempt to summarize these approaches and categorize them according to common characteristics, mainly along the *techniques used* (cf. Section 4.6). Figure 13 shows the distribution of studies over the categories described through the following subsections.

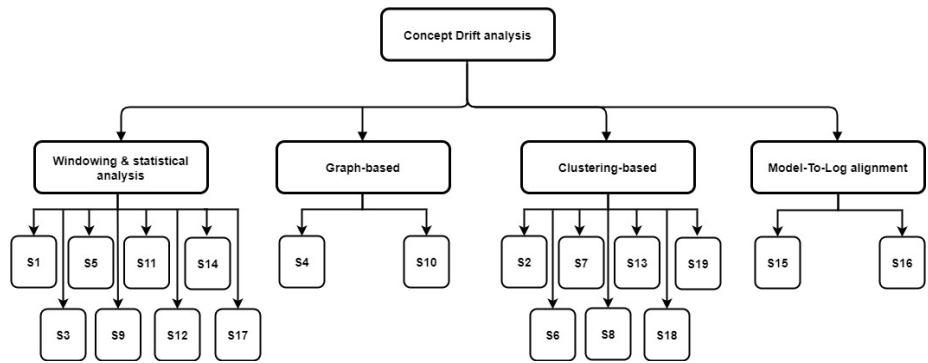

**Figure 13.** Categories of selected studies based on used techniques.

As a primary observation from Figure 13, the biggest share of studies depend either on:

1. windowing to select the traces that shall be considered for drift analysis and for which a statistical analysis is conducted, or on
2. clustering-based techniques to find groups of traces that share similar characteristics that can be generalised and used to detect drifts.

Less studies depend on graph-based analysis techniques to detect drifts or on model-to-log alignment. The lack of studies adopting graph-based analysis techniques can be explained with the computational complexity of graph analysis techniques, as graph matching is NP-complete [47] and some balance is needed between this inherent complexity on the one hand and the accuracy of computations on the other. Note that such balance is missing concerning in the results presented in [5] (cf. Section 5.2).

*5.1. Approaches Relying on Model-to-Log Alignment*

As it will be illustrated in Section 7, some concept drift analysis approaches rely on the presence of a process model along with an event log as inputs to the analysis process. This reliance raises similarities between concept drift and conformance checking tasks. These similarities were considered not to be in the techniques and mechanism. However, they were enough to blur clear lines between goals of both tasks. Early trials into the direction relying on model-to-log alignments are provided by conformance checking, which aims to detect differences between normative and descriptive behaviour. However, conformance checking is not suitable for detecting concept drift as it requires the presence of a precise normative model. Maintaining this normative model, however, becomes costly when the number of traces increases and the cost function of these traces cannot distinguish between non-fitting and almost fitting traces [5].

As an alternative, the authors of [34] suggest the use of *event-based alignment* rather than *trace-based alignment*. In particular, event-based alignment can be achieved through the *vertical* and *horizontal partitioning* of traces as well as the alignment of the resulting sub-logs when checking them against relevant sub-process models. *Vertical partitioning* corresponds to the situation when an event log is divided into sublogs to be processed on several computing machines. In this situation, traces in the original event log are divided based on a selected criterion, for example, case Id or start time of the first event in a trace. Meanwhile, *horizontal partitioning* corresponds to the situation when traces in a given event log are divided based on activities. The authors of [34] argue that partitioning is useful even if all partitions reside on a single computer. Note that the authors of [34] do not provide a numerical analysis of the effects of vertical and horizontal partitioning on computational overhead, but shows that conformance checking that is based on the vertical and horizontal partitioning of an event log can be accomplished easier for declarative models compared to procedural ones.

The authors of [33] employ the same concept of model and trace alignment for comparing variants of the same reference process model across multiple organisations. As main contribution, a visualisation is created of the alignment process over multiple event logs and multiple process models simultaneously. In turn, this is achieved through the construction of an *alignment matrix*, which enables the replay of logs on process variants to decide on the most frequent process variant or the most relevant one.

## 5.2. Graph-Based Analysis Approaches

As a common theme of approaches in this category, visualisations are created as an intermediate step in analysing drifts. The authors of [5] propose an approach for computing mappings between data and time-based partially ordered traces. Time-based partially ordered traces are used to identify dependencies between events. Note that data-based partially ordered traces enable partial similarity rather than complete similarity of traces and can enable localising drift detection to deviating events. Furthermore, partially ordered traces enable a more precise expression of causal dependencies, uncertainty, and co-occurrence of events [48]. First, the data-based partial ordering over events means that two events are dependent if they access the same data attributes [5]. Time-based partial ordering over events, in turn, means that two events are dependent if they have different timestamps, that is, the event with the earlier timestamp is supposed to lead to the occurrence of the event with later timestamp, as they are considered to be concurrent [48].

To compute mappings, the authors of [5] introduce two algorithms. One algorithm uses backtracking with a heuristic function, whereas the other one is a greedy algorithm. Such mapping depends on both local and global similarity measures. On one hand, local similarity between two events is concerned with their properties and local execution contexts which can be measured based on the similarity of neighbours of these events. On the other side, global similarity concerns the similarity of the events' positions in the execution graphs. Defining local and global similarity measures for each event allows providing contextual information about the events, rather than just considering the similarity of an event with another one as in the case of sequential alignment. These mappings are then used to compute *REGs* for traces with similar behaviour as well as relations between the activities. The REGs can then be used to locate and visualise dissimilar behaviour, while giving insights into strongly supported behaviour. The authors of [5] argue that the accuracy of the results obtained from the backtracking mapping approach with data-based partially ordered traces is better than the one of the results obtained when sequential ordering is applied. However, when detecting an increasing number of deviations, the inaccuracy of the presented approach is clear. Finally, when applying the proposed approaches to real-life logs, existing approaches perform better with respect to deviation detection accuracy.

Finally, the authors of [27] introduce an approach dicing an event log based on process perspectives other than control flow. Different versions of the same process can be compared along any perspective by constructing *perspective graphs*. The resulting perspective graphs are then compared leading to *differential perspective graphs*. Although flexible ways for exploring event logs, belonging to the same process are provided, the approach presumes that an event log follows a basic structure that enables the proposed exploration.

## 5.3. Clustering-Based Approaches

Another group of approaches clusters traces in order to detect changing behaviour in an event log. The basic idea is to construct profiles measuring features for each trace, and apply distance measures between instances to be able to cluster them suitably. However, clustering-based approaches have been criticised for their inability to include context information, besides the need to determine the number of clusters beforehand in some clustering algorithms [32]. In turn, the authors of [23] suggest forming *trace profiles of features* like, for example, activity occurrence and resources involved. These profiles can then be used to characterise and differentiate traces as well as to measure the distance between traces based on commonly used distance measures. The authors of [23] ignore the relations and ordering of activity occurrences when forming the profiles, but focuses on activities number of occurrences in differentiating and clustering traces into subsets.

In [32], overcoming the drawbacks of the aforementioned clustering techniques is tried by applying the *Markov Cluster (MCL)* algorithm to event logs. Traces are distinguished through a similarity matrix based on *cosine similarity*. Furthermore, outlier detection and trace clustering are combined, while providing a better illustration of the clustering basis rather than just indicating whether the behaviour is normal or exceptional. Besides, the use of MCL does not require setting the expected number of resulting clusters ahead. Further, it produces a number of clusters of different densities and sizes. To be able to infer the occurrence of a new change, the approach in [32] relies on differences in clusters' sizes and densities to distinguish exceptional from common behaviour. However, these reflections depend on the input, that is, the similarity matrix whose sensitivity to changes depends on the similarity measurement and thresholds used.

By measuring similarity along several perspectives, the approach in [32] utilizes context information rather than only considering control flow similarity. However, the approach does not provide a way to cope with loops and the frequency of events within a trace when constructing the similarity matrix. The ordering of events and their causal relations are not tested along the approach. The absence of such test disables making conclusions about the conditions under which a variation or a change happens. Finally, the approach does not provide an examination of neither the effect of a window size on the resulting clusters nor the efficiency of the proposal in change identification.

The authors of [36] add the time perspective as a factor for clustering the traces of an event log and for using these clusters to discover process variants. The approach starts with forming vectors of *Maximal Repeats (MR)* and adding a time perspective to these vectors that represents the starting time of each trace. It might have been more useful to rely on time similarities concerning single activities, and not just the trace as a whole. Vectors of traces or feature sets are compared using *Euclidean distance*, and clusters are formed using *Agglomerative Hierarchical Clustering (AHC)*. In an experiment, where there are three examples with non-similar processes with a very slight time overlap, Reference [36] emphasizes that using the same approach while considering the time perspective yielded accurate results concerning the discovered model.

A 3-staged approach for analysing concept drift is introduced in [29], as follows:

1. Transforming the event log into a relation matrix based on two relations, namely direct succession and weak order relation. Note that the weak order relation does not provide a reliable characterisation of any trace, as it counts all potential relations that may occur between two activities in a given trace, whether or not these relations already exist.
2. For each relation, a frequency level of values [always/never/sometimes] is determined to localize candidate change points, that is, positions where a change in the relations between two activities is detected. The latter occur when the frequency level for a given relation is changed from one value to another one within a given interval. The length of this interval is restricted by a pre-decided threshold.
3. Candidate change points are combined for the whole event log to gain an overall view of change points in the whole process recorded in the event log.

Although the authors of [29] claim that the proposed approach is able to detect sudden drifts, the change points detected are not precise. Besides, in absence of time-related information, the frequency level of the value "sometimes" might give the impression of gradual drifts as well.

The authors of [4] use recursive partitioning based on conditional inference over event attributes to overcome the limitations of trace clustering techniques. The approach introduces points of interest, which are defined as states or transitions found in a transition system and which are reachable by a portion of the traces adopting this process model. However, the decision on whether or not a certain point of interest is relevant, is defined by a threshold set by the user. The approach then splits traces reaching these points according to relations between dependent and independent event attributes (e.g., time, and resources).

The approach provides flexibility concerning the similarity criteria to be used and the perspectives to be considered, that is, to not only stick to control-flow relations like in trace clustering approaches. However, the approach divides traces based on dependent and independent attributes as well as the degree of their correlation, which is similar to trace clustering despite the number of clusters not being defined ahead. Users need to define a threshold upon which the relevance of a point of interest can be decided. Finally, the user may define the points of interest upon which traces are partitioned. The authors of [27] and [4] have in common that both approaches try to leverage process variants finding to perspectives or attributes other than control-flow differences.

The authors of [8] construct clusters based on similarities between DECLARE constraints. The approach starts with dividing the event log based on a predefined window size and step between windows. Then, DECLARE constraints are discovered from the resulting sublogs. Furthermore, the cases recorded in sublogs are clustered based on homogeneity between the discovered DECLARE constraints in terms of behaviour and time. Afterwards, change points are identified that indicate concept drift positions. Finally, drifts are visualised using *drift maps* (for visualising drifts found in the whole event log) and *drift charts* (for visualising drifts found in a certain cluster). A measure called *ERTC* is introduced in [8] for quantifying the extent of the drift change. ERTC measure is used to define the cluster with the most erratic behaviour, by calculating the Eculidean distance between consectutive values in a time series within the defined window. Note that the erratic behaviour in a cluster is indicated by longer lengths of poly lines in a drift chart, and hence, a higher ERTC value. This approach is able to detect and localise different types of concept drift.

Most approaches presented in this subsection have in common that they start with defining a behavioural profile or a feature set for the activities executed in a process. A profile can be a causal behavioural profile, that is, a representation of relations between every pair of events or activities appearing in an event log [6], a causal footprint, relation set, ordering relations graph, event structures, trace similarity matrix [32], traces' profile [23], feature set [36], or features' vector [13,21]. These profiles are used to capture the behavioural relations or similarities between events based on criteria like number of occurrences or values of associated attributes. Note that most of these profiles have been criticised for neither representing any notion of equivalence nor providing diagnosis of differences between pairs of models [6]. They have been further criticised for their incapability of capturing certain behavioural patterns in a process, for example, task skipping, concurrency, looping, or transitive relations [6]. Some of the presented approaches just measure local properties across a trace, whereas others measure global properties across the entire event log or a subset of the latter. In consequence, a drift is detected either trace or event-based. Accordingly, similarities are then measured either through clustering or through statistical methods.

### 5.4. Approaches Based on Windowing and Statistical Analysis

A common theme which distinguishes this category of approaches in their reliance on statistical approaches to detect changes in event logs divided into windows of either fixed or changeable sizes. The authors of [35] present an approach based on abstract interpretation. It adapts an abstract domain of *convex polyhedra*. A *Parikh vector* is formed representing the number of occurrences of activities in a certain trace. Note that parikh vectors are used for representing features upon which the traces will be classified as belonging to certain polyhedron. Note that this approach is only able to detect the occurrence of a change, but cannot localise it. Furthermore, it depends on the number of occurrences of activities to characterise a trace without regarding other behavioural aspects manifested through the trace (e.g., the relations between its activities). Corresponding experiments were conducted in a fully artificial situation, neither with considering multiple overlapping drifts nor drifts that may not affect the resulting Parikh vectors in an observeable manner (e.g., parallelism of activities or changing ordering relations between activities). This method was criticised by the authors of [21,31] for its inability to detect multiple drifts at the same time, its poor performance [31], and its inability to pinpoint the exact moment of the drift.

In [13], an event log is divided into *time-based windows*. Based on this division, feature vectors are defined either locally for each trace or globally for a subset of the log. The features are used to characterise the current situation of relations between pairs of activities. Furthermore, these features can be also used to detect a change whenever a difference in these relations occur. The authors consider a series of successive populations, whose size is predefined as windows. Then, the differences between two populations are manually investigated. The authors of [31] criticize the need for user interventions by specifying the features used for drift detection as well as the unsuitability of the approach for identifying certain types of drifts, for example, the insertion of conditional branching in a process model.

The authors of [21] modify the approach presented in [13] by using adaptive analysis window size, that is, the number of traces to be compared may be adapted. Furthermore, the authors of [21] propose an approach for *automatic change point detection* through comparing significance values of two populations against a pre-specified threshold. Hypotheses analysis is used to produce these significance values. The basic contribution includes the analysis of gradual drifts as well as the detection and localisation of multi-order dynamics. Experiments prove that runtime and size of steps between windows are inversely proportional [21]. However, increasing step size has a negative effect on the amount of gained information. Although the authors of [21] claim addressing the multi-order dynamics perspective of concept drift, The authors of [21] consider it from both time and granularity sides. Note that the situation would be different if there is an overlap between different types and forms of drifts. The question is whether or not the proposed approach is able to distinguish different drift patterns. In the experiments, the approach in [21] does not include an examination of whether the approach is able to detect and localise multi-order drifts when the length of time intervals between traces arrival is not fixed. Finally, the approach in [21] requires intervention of users with prior knowledge of the type of the drift to be detected; if a gradual drift shall be detected, then the user needs to set a minimum and a maximum gap [9].

An approach that overcomes the limitations observed in [13] and [21] is proposed in [9]. This approach can detect both sudden and gradual drifts from the same event log. This becomes possible without need for user intervention to decide which features shall be captured for log activities or to decide thresholds and gap sizes. The approach starts with dividing the recently observed traces into a window for referencing and another window for detection. Then, it discovers concurrency and causal relations between traces. Afterwards, completed instances are transformed into runs, and hypotheses tests using *Chi-square tests* are conducted on the contingency matrix to obtain the *p*-value. This matrix captures the frequencies of relations between activities of runs in each window. If a number of successive statistical tests have a *p*-value less than a threshold, a drift will be detected.

As a fundamental difference to the other approaches, the approach in [9] encodes gradual drift detection in the form of detecting two sudden drifts. Afterwards, the set of traces after these two sudden drifts are observed to check whether or not they follow a mixture of the distributions before the first and after the second drift. This ensures that the detection process yields a gradual drift rather than two abrupt distinct changes in the process. The authors of [21] captures gradual drifts in the form of non-continuous populations divided by a gap. This way, traces in the first population belong to the process before the drift and traces in the second population belong to the process after the drift took place. The approach in [21] can be criticised for ignoring two issues—*first*, the possibility that there may be two sudden drifts rather than a gradual drift one, and *second*, the co-existence interval where the two process variants co-exist. In turn, the authors of [9] differentiate between sudden drifts on one side and the two sudden drifts limiting an interval of a gradual drift through examining the in-between interval of traces.

The approach proposed in [31] is similar to the one proposed in [9], except for two claimed abilities: detecting drifts within a log that contains traces with high variability and enabling online concept drift detection and localisation. Online concept drift detection and localisation means to detect and locate a drift before completing the trace. The approach is based on statistical tests over the distribution of an abstraction of complete traces. However, the authors of [31] provide no evidence for the efficiency of the approach in the context of online detection. The authors consider the time required to update the concurrency relations as well as to perform the statistical tests as indicators about the suitability of the method for online drift detection, but do not conduct any other evaluation experiments on this issue.

An event log is divided into reference and detection sublogs in [30]. This division takes the form of adaptive windows. Afterwards, process models are discovered for the two sublogs, and graph metrics are generated for them. Used graph metrics include the number of nodes/edges, graph density, in/out degree of each node, and node/edge occurrences. Finally, statistical tests are performed on the generated graph metrics to localise drifts. The authors of [30] claim that statistical tests over the occurrence of egdes yielded the most accurate results. It is argued that detecting and localising concept drift through graph metrics is the basic contribution for being able to localise and characterise drifts more precisely. However, detecting a drift through structural differences between process models does not enable identifying the pattern of the drift.

The authors of [3] provide a comprehensive way to support online process mining as well as to detect various concept drift types. The latter include recurring, incremental, sudden, and gradual drifts. This approach depends on event streams instead of complete event logs to continuously create process models that represent these event streams. However, the approach defines a fixed window size for keeping trace identification. This approach is criticised for its dependency on discovery and conformance checking without considering performance costs, for example, the cost of keeping process histories or checking conformance whenever new events occur. Although the approach to detect drifts is simple, it does not provide distinguishing criteria neither for differentiating types of drifts nor localising a drift.

The authors of [26] propose an extension of the work presented in [3] with the aim to detect data drifts, that is, changes in data attributes associated with events in an event stream. Despite the ability of the proposed approach to detect multiple data drifts in a real-life log from the manufacturing domain, the approach in [26] is unable to differentiate the different types of concept drift under different settings in the evaluation process. As novel contribution, this approach is able to detect drifts in the data perspective in an online setting, that is, from event streams, unlike the work in [28] which accomplishes the same task in an offline setting, that is, from the event log.

## 6. The CONDA-PM Framework

The ultimate purpose of this SLR is to organise knowledge on the topic of concept drift analysis in process mining into a comprehensive structure. Based on this knowledge, we can derive a framework that can be used as a benchmark for developing new approaches to concept drift analysis on one hand, besides being used to compare and evaluate contemporary concept drift approaches on the other. This section introduces the *CONDA-PM (CONcept Drift Analysis in Process Mining) framework* derived by us. It is divided into four phases: goal design, approach coding, implementation, and evaluation. Under the four phases of CONDA-PM, we specify dimensions that may guide users in evaluating the maturity of an approach to concept drift analysis in process mining. The dimensions of CONDA-PM are obtained from the research questions we presented in Section 4.1 with SLR data extraction, as presented in Section 4.6, besides being inspired by the insights gained in Section 4.7. Table 5 shows the dimensions of CONDA-PM and indicates how they are spread over the four phases of the framework.

**Table 5.** CONDA-PM framework.

| The CONDA-PM Framework | | | |
|---|---|---|---|
| **Goals Design** | | **G01:** Defining the granularity level at which the approach operates | |
| | | **G02:** Configuring the handling mode in which a process instance is analysed | |
| **Approach Coding** | **Input Investigation** | **AI1:** Choosing artefacts to be used | |
| | | **AI2:** Analysing inputs to define regarded event log perspectives | |
| | | **AI3:** Defining concept drift type to be addressed | |
| | **Processing** | **AP1:** Concept drift task addressed | |
| | | **AP2:** Defining the change form to be studied | |
| | | **AP3:** Applying suitable techniques | |
| **Implementation** | | **I01:** Approach implementation and tool availability | |
| | | **I02:** Choosing deployment platform | |
| **Evaluation** | | **E01:** Choosing evaluation artefacts | |
| | | **E02:** Evaluation criteria (what to evaluate) | |
| | | **E03:** Used evaluation techniques and metrics (how to evaluate) | |

*6.1. Goals Design Phase*

The ultimate goal of this phase is to define the basic aims of the approach for concept drift analysis in process mining. In this phase, two tasks whose outputs represent important determinants of the approach, need to be accomplished:

- *G01:* Defining the granularity level at which the approach operates. This task is concerned with specifying whether the approach is going to analyse concept drift at the process type level (variability) or process instance level (flexibility). To evaluate the maturity of an approach in this respect, we use a 3-value scale (1, 2, 3) to indicate whether the approach analyses drifts at process level or process instance level or both, respectively. We tend to give higher values to flexibility, as concept drifts are more captured at less granularity levels, that is, the process instance level in this situation. At the process level, changes tend to be more planned, while capturing changes at the process instance level allows reflecting unplanned changes.

- *G02:* Configuring the handling mode in which a process instance is analysed. This task requires specifying whether the approach is going to address concept drift in an online mode (e.g., when many running process instances are under execution), or drift analysis is going to be addressed in an offline mode (i.e., after all process instances are complete and recorded in the event log). To evaluate maturity of an approach in this respect, we use a 2-value scale (1, 2) to indicate whether the approach analyses drifts in offline or onlines mode, respectively. We tend to give higher values to analysing concept drift in an online mode, as the outcomes of conducting online analysis of concept drift might increase robustness, that is, not just responding to change, but also overseeing it and carrying on proactive decision making process. Furthermore, analysing drifts online fosters predictive monitoring and provides more insights into the current status of a running process instance.

*6.2. Approach Coding Phase*

This phase represents the core of any concept drift analysis approach. It is divided into two sub-phases:

### 6.2.1. Input Investigation

In this subphase, the kind of the input to an approach needs to be understood. Based on the insights obtained from the inputs, in turn, the correct technique to be used may be decided. This sub-phase includes the following tasks:

- *AI1:* Choosing artefacts to be used. This task is concerned with specifying the inputs that the approach shall process. To evaluate the maturity of an approach in this respect, we use a 2-value scale (1, 2) to indicate whether the approach uses the event log solely or along with the process model, respectively. We tend to give higher values to an approach which uses a process model along with an event log, because this may yield more comprehensive outcomes, enable studying different forms of change, complement event log chosen perspectives, and give a wider spectrum of techniques to be used.

- *AI2:* Analysing inputs to define regarded event log perspectives. This task is concerned with specifying for which event log perspective the approach shall analyse changes. It is also needed to be decided whether the approach shall focus on one perspective or shall consider multiple event log perspectives. To evaluate the maturity of an approach in this respect, we use a 3-value scale (1, 2, 3) to indicate whether the approach studies only drifts in the control flow perspective or two perspectives or more, respectively.

- *AI3:* Defining the concept drift type to be addressed. This task is concerned with deciding which concept drift category will be addressed (cf. Section 3.2.1). Deciding which category is highly affected by the variance in data available through an event log. To evaluate the maturity of an approach in this respect, we use a 3-value scale (0, 1, 2) to indicate whether the approach does not specify the type of the drift or studies only sudden drifts or studies two or more drift types, respectively.

### 6.2.2. Processing

In this subphase, the inputs are processed according to the chosen concept drift task and change form. This sub-phase includes the following tasks:

- *AP1:* Concept drift task addressed. This task is concerned with deciding whether the approach shall detect, localise, or predict the evolution of a concept drift (i.e., how the business process shall be affected by a combination of drifts, not one drift solely), or enable a combination of these tasks (cf. Section 3.2.2). To evaluate the maturity of an approach in this respect, we use a 4-value scale:

  - *Value of 0* indicates that the focus of an approach is not clear through the study.

  - *Value of 1* indicates that the approach is only able to detect a drift or localise it.

  - *Value of 2* indicates that the approach is concerned with both detection and localisation.

  - *Value of 3* indicates that the approach handles all three tasks of concept drift analysis.

- *AP2:* Defining the change form to be studied. This task is concerned with defining the change form supported by the approach. The decision on this is influenced by the type of inputs as well as the process perspectives covered by the approach. To evaluate the maturity of an approach in this respect, we use a 2-value scale (1, 2) to indicate whether the approach either analyses structural or behavioural changes or the former studies both forms of change, respectively.

  In order to evaluate the maturity of an approach in tasks AI2, AI3, AP1, and AP2, we tend to give higher value to choices which may enable a more comprehensive approach. Note that in AP2, we tend to assign the same value (i.e., 1) to an approach when it analyses either forms of change. This equal assignment is due to the fact that a more comprehensive approach is the one that provides mechanisms to analyse both forms of change. Whereas, analysing any of the two forms of change does not have priority over the other one.

- *AP3:* Applying suitable techniques. This task is concerned with deciding which techniques shall be used to analyse concept drift. Note that the chosen technique depends on the outcomes of the former tasks.The techniques used in the 19 selected studies along with a discussion of their strengths and limitations are presented in Section 5. Furthermore, we list the used techniques as a criterion extracted from the selected studies and used to evaluate these studies in Table 9.

### 6.3. Implementation Phase

Through this phase, an approach is implemented and deployed to a platform that may be used by interested users. Regarding the implementation, we derive two tasks:

- *I01:* Approach implementation and tool availability. This task is concerned with coding an approach and implementing the proposed algorithm. To evaluate the maturity of an approach in this respect, we use a 2-value scale (0, 1) to indicate whether the approach is limited to the theoretical concept described by the respective study or there is an available implementation, respectively.
- *I02:* Choosing deployment platform. This task is concerned with deploying the implementation of an approach and making it available for further usage and evaluation by other researchers. To evaluate the maturity of an approach in this respect, we use a 3-value scale (0, 1, 2) to indicate if there is not an available implementation or there is a prototype or a package is implemented in either ProM or Apromore (https://apromore.org/), respectively. We tend to give higher values to approaches where an implementation is made available through a platform like, for example, ProM or Apromore.

### 6.4. Evaluation Phase

In this phase, the approach is evaluated with respect to the achievement of its design goals. This includes a consideration of how the approach analyses concept drift taking the covered drift tasks and types into account. In this phase, three tasks should be carried out:

- *E01:* Choosing evaluation artefacts. This task is concerned with choosing to evaluate the applicability of an approach on event logs available from either real-life case studies, or artificially generated. To evaluate the maturity of an approach in this respect, we use a 4-value scale (0, 1, 2, 3) to indicate whether no evaluation input is used or the approach is evaluated using only artificial logs or real-life logs or both, respectively.
- *E02:* Evaluation criteria (i.e., what to evaluate). This task is concerned with defining which aspects of the approach with be evaluated. We list which aspects are evaluated as a criterion extracted (if available) from the 19 selected studies and used to evaluate them in Table 10.
- *E03:* Used evaluation techniques and metrics (i.e., how to evaluate). This task is concerned with defining how to apply and evaluate an approach, that is, choosing evaluation techniques and metrics. We list how the approaches in the 19 selected studies are evaluated (if available) as a criterion in Table 10.

Table 6 maps the dimensions of the CONDA-PM framework to the SLR criteria extracted from the selected studies, besides providing a summary of the evaluation scales that will be used for the reminder of this paper to compare and evaluate contemporary approaches in concept drift analysis.

**Table 6.** Mapping CONDA-PM dimensions to systematic literature review (SLR) characterisation criteria.

| Dimension | SLR Criteria | Maturity Evaluation Scale |
|---|---|---|
| G01 | Theme (Variability / Flexibility) | 3-value scale [1–3] |
| G02 | Handling modes (Online/Offline) | 2-value scale [1–2] |
| AI1 | Input (Event log/ Process model) | 2-value scale [1–2] |
| AI2 | Regarded perspectives (Control-flow/Data/Resources/Time) | 3-value scale [1–3] |
| AI3 | Drift patterns (Sudden/Gradual/Recurring/Incremental/ Multi-order) | 3-value scale [0–2] |
| AP1 | Study focus (Detection/ Localisation/ Evolution) | 4-value scale [0–3] |
| AP2 | Change form (Behavioural/ Structural) | 2-value scale [1–2] |
| AP3 | Used techniques | Description & criticism (if available, in Section 5 & Table 9) |
| I01 | Tool availability | 2-value scale [0–1] |
| I02 | Tool availability (platform) | 3-value scale [0–2] |
| E01 | Evaluation input (Artificial / Real-life) | 4-value scale [0–3] |
| E02 | Evaluation (What) | Description & criticism (if available, in Section 5 & Table 10) |
| E03 | Evaluation (How) | Description & criticism (if available, in Section 5 & Table 10) |

## 7. CONDA-PM Application and Evaluation

In order to shed light on research gaps that should be studied in depth, we apply SLR criteria on the selected 19 studies. We divide these criteria over four tables, representing the four phases of CONDA-PM. Table 7, in turn, analyses the existing approaches along the dimensions of the goal design phase (cf. Section 6.1). Moreover, Table 8 analyses the existing approaches along the dimensions of the input investigation sub-phase (cf. Section 6.2), whereas Table 9 analyses these approaches along the dimensions of the processing sub-phase (cf. Section 6.2). Finally, Table 10 analyses the existing approaches along the dimensions of the implementation and evaluation phases (cf. Sections 6.3 and 6.4).

From Table 7 we can make an important observation. S2 claims analysing concept drifts on different granularity levels by visualising drifts occurring in the overall process *(drift maps)* as well as drifts occurring in interesting clusters *(drift charts)*. However, visualising drifts in the overall process can be considered as a study of the evolution of drifts not as an analysis of drifts on different granularity levels. Regarding S12, multi-order drifts are studied (cf. Table 8) as concept drift is considered at different granularity levels (i.e., process type and process instance levels). Therefore, having a comprehensive approach in this regard is still an open research area.

Note that it is crucial to provide an approach measuring whether a drift is qualified to represent a process variant or is a symptom of process flexibility at the process instance level. Having an approach that is able to switch between granularity levels and detect changes at both levels can enable a decision of whether to propagate a change to the process level or to impose preventive and/or corrective actions to deal with the drift consequences.Such studies can provide decision makers with insights on how the business process is changing and the implications of this changes on both behavioural and structural aspects of the process.

As can be observed from Table 8, there is a strong focus on changes in the control-flow perspective, whereas only few studies enable concept drift analysis with respect to the other process perspectives like time or resources. Note that S1 and S3 as indicated in Table 8. S1 and S3 depend on event streams as inputs, as the handling mode of the two approaches (cf. Table 7) is online mode. Consequently, event logs are serialised into event streams without losing information.

**Table 7.** Applying the dimensions of the goal design phase to the selected studies.

| Publication | Theme | | Handling Mode | |
| --- | --- | --- | --- | --- |
| | Variability | Flexibility | Online | Offline |
| S1 | | ✓ | ✓ | |
| S2 | | ✓ | | ✓ |
| S3 | | ✓ | ✓ | |
| S4 | ✓ | | | ✓ |
| S5 | | ✓ | | ✓ |
| S6 & S13 | | ✓ | | ✓ |
| S7 | ✓ | | | ✓ |
| S8 | | ✓ | | ✓ |
| S9 | | ✓ | | ✓ |
| S10 | | ✓ | | ✓ |
| S11 | | ✓ | ✓ | |
| S12 | ✓ | ✓ | | ✓ |
| S14 | | ✓ | | ✓ |
| S15 | ✓ | | | ✓ |
| S16 | | ✓ | | ✓ |
| S17 | | ✓ | ✓ | |
| S18 | ✓ | | | ✓ |
| S19 | | ✓ | | ✓ |

                                                                     

**Table 8.** Applying the dimensions of the input investigation sub-phase to the selected studies .

| Publication | Input | | Regarded Perspectives | | | | Patterns of Change | | | | |
|---|---|---|---|---|---|---|---|---|---|---|---|
| | Event log | Process Model | Control-Flow | Data | Resources | Time | Sudden | Gradual | Recurring | Incremental | Multi-Order |
| S1 | Event Streams | | ✓ | ✓ | | | ✓ | ✓ | ✓ | ✓ | |
| S2 | ✓ | | ✓ | | | | ✓ | ✓ | ✓ | ✓ | |
| S3 | Event Streams | | ✓ | | | | ✓ | ✓ | ✓ | ✓ | |
| S4 | ✓ | | ✓ | ✓ | ✓ | ✓ | | | (NA) | | |
| S5 | ✓ | | ✓ | | | | ✓ | ✓ | | | |
| S6 & S13 | ✓ | | ✓ | ✓ | | ✓ | | | (NA) | | |
| S7 | ✓ | | ✓ | ✓ | ✓ | ✓ | | | (NA) | | |
| S8 | ✓ | | ✓ | | | | ✓ | | | | |
| S9 | ✓ | ✓ | ✓ | | | | | | (NA) | | |
| S10 | ✓ | | ✓ | | | | | | (NA) | | |
| S11 | ✓ | | ✓ | | | | | | (NA) | | |
| S12 | ✓ | | ✓ | | | ✓ | ✓ | ✓ | | | ✓ |
| S14 | ✓ | | ✓ | | | | ✓ | ✓ | | | |
| S15 | ✓ | ✓ | ✓ | | | | | | (NA) | | |
| S16 | ✓ | ✓ | ✓ | | | | | | (NA) | | |
| S17 | ✓ | | ✓ | | | | | ✓ | | | |
| S18 | ✓ | | ✓ | | | ✓ | | | (NA) | | |
| S19 | ✓ | | ✓ | | | | | | (NA) | | |

**Table 9.** Applying the dimensions of the processing sub-phase to the selected studies.

| Publication | Study Focus | | | Change Form | | Used Techniques |
|---|---|---|---|---|---|---|
| | Detection | Localisation | Evolution | Behavioural | Structural | |
| S1 | ✓ | ✓ | | ✓ | | Sliding window, Statistical Analysis for outlier detection & Serialisation in YAML |
| S2 | ✓ | ✓ | ✓ | ✓ | | Sliding window, Hierarchical Clustering, Change point detection using PELT & Declarative constraints discovery using MINERFUL |
| S3 | ✓ | | | ✓ | | Sliding window, Stream–Based Abstract Representation (SBAR), Hashtables, Inductive Miner |
| S4 | | (NA) | | ✓ | | Perspective graphs & finding statistical differences between them to result in differential perspective graphs |
| S5 | ✓ | ✓ | ✓ | ✓ | | Adaptive windowing, statistical testing (Chi–Square test) & oscillation filter |
| S6 & S13 | ✓ | | | ✓ | | MCL & cosine similarity |
| S7 | | (NA) | | ✓ | | Recursive Partitioning by Conditional Inference (RPCI), points of interest & event augmentation |
| S8 | ✓ | | | ✓ | | DBSCAN clustering |
| S9 | ✓ | ✓ | | | ✓ | Adaptive windowing, Heuristic mining & G-Tests |
| S10 | ✓ | ✓ | | ✓ | ✓ | Backtracking, heuristic function, a greedy algorithm, cost function & REGs |
| S11 | ✓ | ✓ | | ✓ | | Adaptive windowing, statistical testing (G–test of independence) |
| S12 | ✓ | ✓ | ✓ | ✓ | | Adaptation of ADWIN approach |
| S14 | ✓ | ✓ | ✓ | ✓ | | Statistical hypotheses testing using Kolmogorov–Smirnov (KS) test, Mann–Whitney test, Hotelling T2 test & windowing |
| S15 | ✓ | ✓ | | ✓ | | Alignment matrix |
| S16 | | ✓ | | ✓ | | Vertical & horizontal partitioning |
| S17 | ✓ | Proposed Ideas | | ✓ | | Abstract interpretation, Parikh vectors & Adaptive windowing |
| S18 | ✓ | | | ✓ | | Euclidean Distance & Agglomerative Hierarchical Clustering (AHC) |
| S19 | ✓ | | | ✓ | | Trace Clustering / divide & conquer (trace profiles) |

**Table 10.** Applying the dimensions of the implementation and evaluation phases to the selected studies.

| Publication | Tool Availability | Evaluation Input | | Evaluation (What/How) |
|---|---|---|---|---|
| | | Artificial | Real-life | |
| S1 | prototype | | log from the manufacturing domain with 10 process instances & 40436 events | the ability of the approach to identify drifts with different thresholds for outlier detection |
| S2 | Python implementation of the algorithms | four synthetic logs | Italian IT help desk log & BPIC'11 log | **what:** the ability to identify different drift types & plotting drifts as charts & maps **How:** comparing the detection results to those in Reference [31] and computing F-scores |
| S3 | a tool to synthesis process execution logs +a web service to transform static logs into event streams + a web service to run algorithms on events | Insurance process augmented by cases representing all types of drifts | | The ability of the proposed algorithms to detect the 4 types of drifts |
| S4 | Multi Perspective Process Comparator MPC plugin in ProM | | BPIC13 & BPIC15 –BPIC13 (cases of an IT incident handling process at Volvo Belgium–IT teams are organized into tech–wide functions, organization lines & countries | **Compared to ProcessComparator plugin in ProM regarding:** differences at the event level (intra–event & inter–event relations) –differences at the fragment level (intra–fragment & inter–fragment graphs)–time–wise differences compared to case–wise differences using a sliding window of 3 days. |
| S5 | ProDrift as a plugin in Apromore platform | loan application (72 event logs [four for each of 18 change patterns]–15 activities in the base model)+18 logs for gradual drifts | motor insurance claims(4509 cases–29108 events) –motor insurance claims of different insurance brand (2577 cases–17474 events) | **Accuracy assessment using:** F–score & mean delay–expert validation of results on real-life logs **What:** impact of window size on accuracy –impact of oscillation filter size –impact of adaptive window –accuracy per change pattern –time performance–comparison with results in Reference [13] |
| S6 & S13 | Trace Clustering package in ProM | 164 variants–1000 cases–17 activities –6812 events | two logs (Dutch hospital–1143 cases–624 activities–150291 events)+ (municipality–1199 cases–398 activities–52217 events) | **What:** Comparison to Trace Alignment& ActiTrac **How:** mining the model for each technique using Flexible heuristics Miner (FHM)–fitness is calculated using cost–based log alignment, cluster entropy& split rate for each technique. |

| Publi–Cation | Tool Availability | Evaluation Input | | Evaluation (What/How) |
|---|---|---|---|---|
| | | Artificial | Real-life | |
| S7 | Process variant finder plugin in Variant–Finder package in ProM | 10000 cases–12721 events–31 resources | Spanish telecommunications company–8296 cases –40965 events representing 5 activities | **For the artificial log:** performance variability is evaluated by choosing time as the dependent attribute& resources as the independent one **For the real-life log:** examining the ability to define variants & evaluating the impact of some of the detected partitions on the whole process |
| S8 | Python implementation of the algorithms | 32 logs of loan application | | –the impact of different thresholds on F-score –the effect of different DBSCAN parameters –comparison to the approach proposed in Reference [49] concerning: time, f-score, when high precision is required, and average detection error. |
| S9 | plugin in ProM | 72 logs from a base model with 15 activities–18 change patterns are applied | | –Accuracy is measured using F-Score –Average delay between actual & detected drifts –comparison to the approaches proposed in Reference [13] &[49] concerning: F-score & average delay |
| S10 | ✓ | 1000 cases –6590 events | 2 logs –Maastricht University medical centre (2838 cases& 28163 events) +governmental municipality (1434 cases& 8577 events) | **5 different experiments:** Accuracy score is calculated to indicate how accurately deviating events are detected, in comparison to 3 different approaches **Experiments are concerned with:** effect of different configurations–effect of using sequential orders instead of partial orders –effect of various deviation levels–performance & scalability–effect of using synthetic& real–life data |
| S11 | ProDrift 2.0 as a plugin in Apromore platform | 90 logs of sizes 2500,5000,10000 cases using a model with 28 activities in CPN | BPIC 2011 log –Dutch academic hospital–1143 cases–150291 events | **Accuracy assessment:** F–score& mean delay. **What:** impact of oscillation filter size–impact of inter drift distance–comparison to previous method they proposed concerning different patterns detection & with different log variability–execution time |

**Table 10.** *Cont.*

| Publi–Cation | Tool Availability | Evaluation Input | | Evaluation (What/How) |
|---|---|---|---|---|
| | | **Artificial** | **Real-life** | |
| S12 | ConceptDrift plugin in ProM | Insurance claim (sudden & multi–order changes: 6000 cases –15 activities –58783 events)–(gradual drifts: 2000 cases–15 activities–19346 events) | Dutch municipality –184 cases –38 activities–4391 events | **What:** measuring the ability to detect & localise changes with a lag window of different traces no. **How:** adaptive windowing using the Kolmogorov test over the data stream then classic data mining metrics are derived. -For the multi–order drifts & gradual drifts: different configurations using different population sizes, step sizes, *p*-value threshold & gap sizes –examples of linear & exponential gradual drifts |
| S14 | Concept Drift plugin in ProM | insurance claim (6000 cases –15 activities –58783 events) | municipality (116 cases –25 activities–2335 events) | The ability to detect (using RC feature) & localize change points (using WC feature) –influence of population size –Time complexity for feature extraction & hypothesis test analysis |
| S15 | Comparison framework package in ProM | | CoSeLoG project –5 municipalities | **What:** Comparison visualization understandability **How:** Meetings with representatives of municipalities- comparison of replay fitness scores & visualization |
| S16 | (NA) | Illustrative examples | | (NA) |
| S17 | (NA) | 13 models are used to derive logs with drifts | 1050 cases–35 activities | Concatenation of two logs with different distributions & measuring the approach's ability to detect the change, with estimation of the number of points needed to sample for change detection. |
| S18 | (NA) | 3 logs–each with 2000 cases–over 1 year | | **What:** the accuracy at which each approach is able to classify different traces. **How:**3 approaches (without time factor –with time=no. of elements in the feature set –with time is weighted against the no. of elements in the feature set) were applied on the 3 logs –an accuracy metric is calculated. |
| S19 | Trace Clustering plugin in ProM | | AMC hospital in Amsterdam (619 cases–52 activities–3574 events) | Different combinations of distance measures & clustering techniques–the understandability of resulting models with the aid of domain experts |

From Table 9 we may conclude that structural changes are not comprehensively studied, whereas focus is put on studying behavioural changes of a process. This can be explained with the challenges facing process model similarity. Representing same behaviour with different structures, differences in labeling styles, and terminology differences all illustrate these challenges [50]. A study considering behavioural and structural changes together may provide a comprehensive approach to explain the semantics of a change and might allow studying the contextual influence on a changing business process.

Concerning the techniques used by each approach, we find that these techniques primarily depend on the kind of input, considered process perspectives, and the change front studied. Note that existing techniques mostly address drift analysis rather than reasoning about the drift. This supports our conclusion that there is not enough research on the reasons of drifts and their context. However, note that studies on the evolution of drifts and the way these drifts contribute to the change of the business process use sliding windows as a technique to partition the event log. Further, they then apply a statistical method to study the differences in distributions between multiple windows of the event log.

Table 10 attempts to provide more insights into the contributions and outputs of the selected studies. We checked whether there exist implementations of the proposed approaches, which we consider as both an indicator of the technical feasibility of an approach and its further usage. Most implementations are made available as plugins or packages in ProM. Despite the availability of Apromore as an open source business process analytics platform, only the implementations of approaches in S5 and S11 are integrated into Apromore.

Other important dimensions include the evaluation inputs and criteria of each proposed approach. As stated before, most of the collected studies use real-life logs for evaluating different aspects of the approaches. However, most studies fail in demonstrating the performance costs of applying the proposed approaches, for example, the time consumed for running the tool as well as the effects of log size or studied drift patterns or handling modes on different performance measurements. There is a lack of studies on time, performance, memory and other costs related to tasks associated with concept drift analysis, for example, detection, localisation, and evolution analysis.

In order to evaluate the maturity of the selected studies along the scales defined for each dimension of CONDA-PM, we constructed Table 11. This table indicates maturity in terms of numbers and categorises the SLR criteria along the four phases of CONDA-PM. Concerning the *goal design* phase, we notice that S1, S3, S11, S12 and S17 have the highest scores regarding the defined maturity scales, whereas for the *input investigation* sub-phase, S1 and S12 show the highest maturity in providing approaches that reflect a higher understanding of the inputs in hand. Regarding the *processing* sub-phase, S2, S5, S10, S12 and S14 show a high maturity level regarding the comprehensiveness of manipulating inputs to reach the defined goals. Note that this maturity scale is complemented by the analysis provided in Section 5 and the description of techniques available in Table 9. Finally, S2, S5, S12, and S14 satisfy the study focus dimension.

We observed a relation between the absence of clearly specifying a study focus and regarding variability as a theme of the study in S14 and S15. Concerning the *implementation* phase, S5, S11, and S12 are dropped down on the maturity scale as no implementation is available for these approaches. Regarding the *evaluation* phase, approaches that use artificial logs side-by-side with real-life logs have the highest scores. These findings are complemented with the descriptions of the evaluation process (cf. Table 10). Although the number of studies is rather low, each one is providing a novel approach, meaning that the topic deserves more in-depth exploration. Still, there is potential to develop complementary approaches to a solution of concept drift analysis in process mining.

**Table 11.** Maturity evaluation of the selected studies along the CONDA-PM dimensions.

| CONDA-PM Criteria | SLR Criteria | Publication | | | | | | | | | | | | | | | | | | |
|---|---|---|---|---|---|---|---|---|---|---|---|---|---|---|---|---|---|---|---|---|
| | | S1 | S2 | S3 | S4 | S5 | S6 & S13 | S7 | S8 | S9 | S10 | S11 | S12 | S14 | S15 | S16 | S17 | S18 | S19 |
| **Phase I: Goal Design** | | | | | | | | | | | | | | | | | | | |
| G01 | Theme | 2 | 2 | 2 | 1 | 2 | 2 | 1 | 2 | 2 | 2 | 2 | 3 | 2 | 1 | 2 | 2 | 1 | 2 |
| G02 | Handling mode | 2 | 1 | 2 | 1 | 1 | 1 | 1 | 1 | 1 | 1 | 2 | 1 | 1 | 1 | 1 | 2 | 1 | 1 |
| **Phase II: Approach Coding (Input investigation)** | | | | | | | | | | | | | | | | | | | |
| AI1 | Input | 1 | 1 | 1 | 1 | 1 | 1 | 1 | 1 | 2 | 1 | 1 | 1 | 1 | 2 | 2 | 1 | 1 | 1 |
| AI2 | Regarded perspective | 2 | 1 | 1 | 3 | 1 | 3 | 3 | 1 | 1 | 1 | 1 | 2 | 1 | 1 | 1 | 1 | 2 | 1 |
| AI3 | Patterns of change | 2 | 2 | 2 | 0 | 2 | 0 | 0 | 1 | 0 | 0 | 0 | 2 | 2 | 0 | 0 | 1 | 0 | 0 |
| **Phase II: Approach Coding (Processing)** | | | | | | | | | | | | | | | | | | | |
| AP1 | Study focus | 2 | 3 | 1 | 0 | 3 | 1 | 0 | 1 | 2 | 2 | 2 | 3 | 3 | 2 | 1 | 1 | 1 | 1 |
| AP2 | Change form | 1 | 1 | 1 | 1 | 1 | 1 | 1 | 1 | 1 | 2 | 1 | 1 | 1 | 1 | 1 | 1 | 1 | 1 |
| AP3 | Used techniques | | | | | | Described in Table 9 | | | | | | | | | | | | |
| **Phase III: Implementation** | | | | | | | | | | | | | | | | | | | |
| I01 | Tool availability | 1 | 1 | 1 | 1 | 1 | 1 | 1 | 1 | 1 | 1 | 1 | 1 | 1 | 1 | 0 | 0 | 0 | 1 |
| I02 | Tool availability (platform) | 1 | 1 | 1 | 2 | 2 | 2 | 2 | 1 | 2 | 1 | 2 | 2 | 2 | 2 | 0 | 0 | 0 | 2 |
| **Phase V: Evaluation** | | | | | | | | | | | | | | | | | | | |
| E01 | Evaluation input | 2 | 3 | 1 | 2 | 3 | 3 | 3 | 1 | 1 | 3 | 3 | 3 | 3 | 2 | 1 | 3 | 1 | 2 |
| E02 | Evaluation (what) | | | | | | Described in Table 10 | | | | | | | | | | | | |
| E03 | Evaluation (How) | | | | | | Described in Table 10 | | | | | | | | | | | | |

## 8. Discussion

It is crucial to have a clear understanding of the stages and targets that should be satisfied by an approach to concept drift analysis. This understanding is shaped in dimensions of the CONDA-PM framework. A comprehensive approach to concept drift analysis should have a goals definition phase, a coding and processing phase, an implementation phase and an evaluation phase. In the goals definition phase, the granularity level and the mode the approach is addressing are identified. In the approach coding and processing phase, the drift patterns to be studied as well as the desired input and the perspectives to addressed are identified. As shown, there is a relation between the inputs and the decision about the covered perspectives. Based on the former decisions, the authors make choices of the used techniques and the change form. In the same phase, the study focus is decided. The implementation and evaluation phases demonstrate the outputs of an approach and show the maturity level reached by the approach.

The results obtained after conducting the SLR, constructing the CONDA-PM framework, and applying it on 19 selected studies highlight some notes. *First,* although concept drift is mentioned as a challenge in the process mining manifesto [1], less studies exist that address this challenge. Compared to other process mining topics, like conformance checking and predictive monitoring, more research efforts should be spent on concept drift analysis. *Second,* to the best of our knowledge, no study exists that reasons about the detected drifts and explains them by contextual analysis neither of the process instance context nor the entire business process context. *Third,* most studies detect and localise concept drifts in offline mode, whereas only few studies enable an runtime online analysis of running processes as well. This research gap disables the use of concept drift analysis approaches in other process mining tasks. As example of missed opportunities for concept drift analysis, consider predictions on a running business process when the data used by the predictions is changing. *Fourth,* there is a lack of studies regarding the effect of a drift on process performance measurements, and the causality between a certain type of a drift (i.e., whether it is a sudden, incremental, recurring or gradual drift) and different performance indicators.

## 9. Related Work

In this paper, we are concerned with characteristing concept drift analysis approaches, that is, the input an approach needs, the analysis task, change form, and drift type the approach is concerned with, and the scale and timing at which the drift is analysed. We are further concerned with providing a framework that defines the basic requirements of a concept drift analysis approach. To the best of our knowledge, no other SLR conducted on concept drift analysis in process mining exits. However, there are some literature reviews that were conducted on related topics in process mining as well as related research areas. In addition, there are related fields that we consider complementary to gain an understanding on how changes and variances in a business process are studied.

In the field of change mining, corresponding approaches are more concerned with change discovery and cartography, rather than detecting or predicting a change. A change log generated by a PAIS like ADEPT or CBRFlow [18], is taken as input to be analysed along with the execution log. The former log [18] captures the process changes performed on some process instances. A change log instance is a sequence of process changes performed on a process instance. Ideally, change logs keep information about high-level changes (i.e., adding, inserting, deleting, or moving events) performed, instead of low-level change primitives (e.g., addNode, deleteEdge). As a remarkable research effort in this direction, Reference [51] evolves the idea of change mining processes to produce a change tree including the number of occurrences of a certain change sequence. This helps predicting how a change sequence would evolve and whether a change deserves to be leveraged to the level of the process type rather than the process instance level. The authors of [52] provide an approach using change trees with the possibility to base the analysis of change sequences not only on the similarity between labels, but also on contextual similarities, like the usage of the same resources or comparable execution times.

A related review is presented in [16], which surveys deviance mining approaches based on sequence mining. The survey summarizes 48 papers and classifies them. From this, five basic approaches are identified. The selected approaches address event log abstraction based on the occurrence count of individual activities or frequent set of trace attributes or features of sequence of events occurring multiple times in a process instance or across different instances. The authors of [16] study seven feature sets derived from the five selected approaches, and evaluates them using five real-life event logs. Fisher score is used for computing weights of statistical correlation of a feature with the outcome classes. Afterwards, a model is constructed using a variety of methods including decision trees, neural networks, and K-Nearest Neighbours (KNN). Finally, classification accuracy and the interestingness of rules extracted to explain the reasons of the deviance are assessed.

In turn, the authors of [53] provide a comprehensive survey of business process variability and provides a comprehensive framework for evaluating process variability approaches. The former is derived based on analysing 63 collected studies from literature. The purpose of the resulting *VIVACE* framework is to analyse how different process variability approaches support process variability through a business process lifecycle. This evaluation is based on the extent to which each evaluated approach realizes a set of features in each BPM stage. *VIVACE* captures the business process perspectives covered by a process variability approach, the variability-specific language constructs are provided for specifying variability, the existence of a tool implementing the approach, the type of empirical evaluation performed to test the approach, and the domain in which the analysed approach has been applied. *VIVACE* is applied to three process variability approaches.

A highly cited study in the field of data mining provides a review and taxonomy of adaptive learning algorithms at the presence of concept drifts [12]. The focus of this study is on reviewing adaptive learning methods for online supervised learning, while real concept drift takes place. According to [12], a real concept drift is *a change in the conditional distribution of the output given the input, while no change happened in the distribution of the input*. Adaptive learning algorithms differ in four fundamental criteria [12]: memory utilisation (including data management and forgetting mechanisms), change detection techniques, learning stages, and loss estimation techniques.

Another important review is presented in [54], where 20 approaches of clustering process instances are studied. These approaches are studied according to their objective, process instance representation applied, distance measure used, category of the clustering approach (i.e., partitioning, hierarchical, density, or depending on neural networks), and tool availability. The authors evaluate these approaches based on two scenarios: the first one concentrates on the ability of an approach to identify and separate process instances belonging to different business processes. The second one concentrates on the ability of an approach to reduce the complexity of the mined process model. The authors use metrics for the latter scenario to quantify the amount of nodes, edges and relations between each other in generated process models.

A framework for classifying and analysing contemporary approaches in predictive process monitoring is presented in [55]. The framework consists of multiple dimensions that characterize predictive monitoring approaches, as well as highlight research points in this field. The framework dimensions include prediction perspective (what to be predicted through the approach proposed by a study), inputs to the approach, algorithms and techniques used, and availability of a tool implementing the approach. The framework is applied to 55 studies. This SLR categorized predictive monitoring-related perspectives into categorical and numerical predictions, besides predictions of the respective next activity during the execution of process instances.

## 10. Summary and Outlook

This article presents an attempt to categorize and characterize studies dealing with concept drift analysis as a topic associated with process mining. Further, it provides a benchmark for assessing the comprehensiveness of approaches dealing with concept drift analysis in process mining. We carefully define all aspects associated with concept drift. These aspects resolve ambiguities resulting from confusing the meaning of concept drift with other idioms like noise as an undesirable behaviour present in an event log. We highlight insights into relations between information available through inputs, studied change forms, concept drift tasks, and selected techniques. The criteria we used to study available approaches highlight shortage in solutions for detecting different types of drifts other than sudden drifts, and different perspectives other than control flow. The SLR highlight a shortage of research in online analysis of concept drifts. The shortage of platforms for hosting implementations of currently available approaches should be addressed. Furthermore, more research effort should be spent on the direction of measuring the performance of concept drift analysis approaches to ensure their efficiency and scalability. CONDA-PM, which is a framework capturing the dimensions of a concept drift analysis approach through four phases, provides a roadmap for developing and assessing upcoming approaches that will be introduced to address concept drift analysis.

**Author Contributions:** G.E. and M.R. drafted and revised the manuscript. M.A. read and revised the manuscript. A.M.R. and S.I.B. revised the manuscript. G.E. and M.R. were responsible for conceptualizing this work. G.E. was responsible for the formal analysis, methodology, and writing the original draft. M.R. was responsible for reviewing and editing this article. M.A. and M.R. validated and supervised this work, while S.I.B. and A.M.R. were responsible for funding acquisition. All authors have read and agreed to the published version of the manuscript.

**Funding:** This study is carried out through fund provided by the Egyptian Ministry of Higher Education and Mansoura University to the first author. This fund is in the course of a joint PhD supervision program between Mansoura University and Ulm University.

**Conflicts of Interest:** The authors declare no conflict of interests.

## Glossary

| | |
|---|---|
| CONDA-PM | CONcept Drift Analysis in Process Mining Framework. A four-staged framework providing guidance on the fundamental components of a concept drift analysis approach in the context of process mining |
| SLR | Systematic Literature Review. A survey of a topic conducted according to systematic steps and adopts a certain format |
| concept drift | the situation in which the process is changing while being analysed |
| normative process model | an ideal view on how business process activities shall be performed and used for monitoring and inspecting how running process instances are executed |
| descriptive model | a process model which captures how process activities are carried out |
| An event log | a record of the actual execution of a process in terms of cases |
| process instance | notion of process executions at the conceptual view of a business process |
| case | notion of process executions at the data view of a business process |
| trace | notion of process executions at the logical view of a business process |
| An event stream | a collection of events representing an incomplete running process instance |
| Control-flow perspective | order of activities executed in a business process |
| Organisational perspective | organisational resources and roles involved in the execution of events |
| Time perspective | the point in time events occur |
| Case perspective | properties of process instances as they are stored in an event log |
| A process history | a list of process models discovered for a business process whenever a violation is detected |
| A process variant | an execution path of a business process which results in different process instances which share some commonalities like sharing some same activities |
| Behavioural profiles | a representation of relations between every pair of events or activities appearing in an event log |
| An execution graph | a directed acyclic graph used to represent the execution of a process instance in terms of nodes (i.e., events), edges (i.e. relations between the nodes), and functions assigning each event to its activity class |
| Procedural process models | a conceptual view of how activities should be carried out |
| declarative process models | a formalisation of the undesired behaviour through defining a set of constraints. In these models, the order of activities execution is not rigid |
| noise | rare and infrequent behaviour which may appear in an event log |
| Deviations | additional behaviour observed in the event log, but not foreseen in the normative process |
| deviance mining | a family of process mining techniques aimed at analyzing event logs in order to explain the reasons why a business process deviates from its normal or expected execution |
| business process drift detection | a family of techniques to analyse event logs or event streams generated during the execution of a business process in order to detect points in time when the behaviour of recent executions of the process differs significantly from that of older cases |
| change log | a log created and maintained by adaptive PAISs, and contains information about process changes at both type and instance level |
| variability | the process of providing a customisable process model |
| flexibility | regards the changing circumstances and variations at the process instance level |
| sudden concept drift | a drift happening suddenly at a certain point during the execution of a process instance |
| Incremental concept drift | a drift is introduced incrementally into the running process until the process reaches the desired version |
| Gradual concept drift | a drift type when two versions of the same process co-exist and process instances following both versions are running, till a certain point when only the new process model is adopted |
| Recurring concept drift | a drift caused by the process context or environment and takes place only over a defined period of time |
| multi-order drift | a change taking place on a macro level involving two different process model versions of the same process and on a micro level representing a slight change in the same process model |
| Concept drift analysis tasks | concept drift detection, concept drift localisation, and concept drift evolution analysis |
| flexibility by change | a condition where a certain change is applied to selected traces at runtime |
| variability by restriction | provides a process model with all allowed behavior and restricting models that may be configured from this model by adding more behavior |
| variability by extension | provides a process model whose behaviour can be extended by deriving other process models based on the behaviour explained by it |
| Vertical partitioning | the situation when traces in an event log are divided based on a selected criterion, e.g., case Id or start time of the first event in a trace and they are processed on several computing machines |
| horizontal partitioning | the situation when traces in an event log are divided based on activities |
| alignment matrix | a matrix used in log-to-model alignment to enable replaying logs on process variants to decide on the most frequent process variant or the most relevant one |
| data-based partial ordering | two events are dependent if they access the same data attributes |
| Time-based partial ordering | two events are dependent if they have different timestamps |
| ERTC | a measure used for quantifying the extent of the drift change |
| VIVACE | a framework which analyses how different process variability approaches support process variability through a business process lifecycle |